\documentclass[lettersize,journal]{IEEEtran}
\usepackage{amsmath,amsfonts}
\usepackage{algorithmic}
\usepackage{algorithm}
\usepackage{array}
\usepackage{multirow}
\usepackage{placeins}
\usepackage{subcaption}
\usepackage{textcomp}
\usepackage{stfloats}
\usepackage{url}
\usepackage{verbatim}
\usepackage{graphicx}
\usepackage{cite}
\usepackage{listings}
\usepackage{xcolor}
\usepackage[normalem]{ulem}
\graphicspath{{figs_eps/}} %

\definecolor{codegreen}{rgb}{0,0.6,0}
\definecolor{codegray}{rgb}{0.5,0.5,0.5}
\definecolor{codepurple}{rgb}{0.58,0,0.82}
\definecolor{backcolour}{rgb}{1,1,1}

\lstdefinestyle{mystyle}{
    escapechar={|}, 
    backgroundcolor=\color{backcolour},   
    commentstyle=\color{codegreen},
    keywordstyle=\color{magenta},
    numberstyle=\tiny\color{codegray},
    stringstyle=\color{codepurple},
    basicstyle=\ttfamily\footnotesize,
    breakatwhitespace=false,         
    breaklines=true,                 
    captionpos=b,                    
    keepspaces=true,                 
    numbers=left,                    
    numbersep=5pt,                  
    showspaces=false,                
    showstringspaces=false,
    showtabs=false,                  
    tabsize=2
}

\usepackage{tikz}
\usepackage{textcomp}
\usepackage[doipre={DOI:~}]{uri}
\usepackage{lipsum}
\newcommand\copyrighttext{%
  \footnotesize \textcopyright 2023 IEEE. Personal use is permitted, but republication/redistribution requires IEEE permission. See https://www.ieee.org/publications/rights/index.html for more information.
  }
\newcommand{\copyrightnotice}{%
\begin{tikzpicture}[remember picture,overlay]
\node[anchor=south,yshift=10pt] at (current page.south) {\fbox{\parbox{\dimexpr\textwidth-\fboxsep-\fboxrule\relax}{\copyrighttext}}};
\end{tikzpicture}%
}
\usepackage{fancyhdr}

\begin{document}
\bstctlcite{IEEEexample:BSTcontrol}

\title{Dynamic Decision Tree Ensembles for Energy-Efficient Inference on IoT Edge Nodes}

\author{%
        Francesco~Daghero,~\IEEEmembership{Member,~IEEE,}
        Alessio~Burrello,~\IEEEmembership{Member,~IEEE,}
        Enrico Macii,~\IEEEmembership{Fellow,~IEEE,}
        Paolo Montuschi,~\IEEEmembership{Fellow,~IEEE,}
        Massimo Poncino,~\IEEEmembership{Fellow,~IEEE,}
        and~Daniele Jahier Pagliari~\IEEEmembership{Member,~IEEE}%
\thanks{Manuscript received January XX, XXXX; revised January XX, XXXX. (\textit{Corresponding Author: Daniele Jahier Pagliari).}}
\thanks{F. Daghero, P. Montuschi, M. Poncino and D. Jahier Pagliari are with the Department
of Control and Computer Engineering, Politecnico di Torino, Turin, 10129, Italy (e-mail: francesco.daghero@polito.it; paolo.montuschi@polito.it; massimo.poncino@polito.it; daniele.jahier@polito.it).}%
\thanks{A. Burrello and E. Macii are with the Interuniversity Department of Regional and Urban Studies and Planning, Politecnico di Torino, Turin, 10129, Italy (e-mail: alessio.burrello@polito.it; enrico.macii@polito.it).}%
\thanks{A. Burrello is also with the Department of Electrical, Electronic and Information Engineering, University of Bologna, 40136 Bologna, Italy (e-mail: alessio.burrello@unibo.it).}%
\thanks{Copyright (c) 20xx IEEE. Personal use of this material is permitted. However, permission to use this material for any other purposes must be obtained from the IEEE by sending a request to pubs-permissions@ieee.org.}}%
\markboth{Journal XXXX,~Vol.~XX, No.~XX, Month~XXXX}%
{Daghero \MakeLowercase{\textit{et al.}}: Dynamic Decision Tree Ensembles for Energy-Efficient Inference on IoT Edge Nodes}

\maketitle
\copyrightnotice

\thispagestyle{fancy}
\fancyhead{}
\fancyhead[C]{This article has been accepted for publication in IEEE Internet of Things Journal. This is the author's version which has not been fully edited and content may change prior to final publication. Citation information: DOI 10.1109/JIOT.2023.3286276}

\begin{abstract}
With the increasing popularity of Internet of Things (IoT) devices, there is a growing need for energy-efficient Machine Learning (ML) models that can run on constrained edge nodes. Decision tree ensembles, such as Random Forests (RFs) and Gradient Boosting (GBTs), are particularly suited for this task, given their relatively low complexity compared to other alternatives.
However, their inference time and energy costs are still significant for edge hardware. 

Given that said costs grow linearly with the ensemble size, this paper proposes the use of \textit{dynamic ensembles}, that adjust the number of executed trees based both on a latency/energy target and on the complexity of the processed input, to trade-off computational cost and accuracy. We focus on deploying these algorithms on multi-core low-power IoT devices, designing a tool that automatically converts a Python ensemble into optimized C code, and exploring several optimizations that account for the available parallelism and memory hierarchy.
We extensively benchmark both static and dynamic RFs and GBTs on three state-of-the-art IoT-relevant datasets, using an 8-core ultra-low-power System-on-Chip (SoC), GAP8, as the target platform. Thanks to the proposed early-stopping mechanisms, we achieve an energy reduction of up to 37.9\% with respect to static GBTs (8.82 uJ vs 14.20 uJ per inference) and 41.7\% with respect to static RFs (2.86 uJ vs 4.90 uJ per inference), without losing accuracy compared to the static model.

\end{abstract}


\section{Introduction}\label{sec:introduction}
\IEEEPARstart{M}achine Learning (ML) inference is increasingly present in multiple Internet of Things (IoT) applications, ranging from human activity recognition~\cite{daghero2021ultracompact} to predictive maintenance~\cite{burrelloPredictingHardDisk2021} or to seizure detection~\cite{manzouri2018comparison}.
A cloud-centric paradigm is traditionally leveraged, with the IoT nodes collecting data, and offloading almost all computations to high-end servers.
This approach allows relying on robust and accurate models, independently from the IoT device's computing capabilities.
Nevertheless, the need for a constant connection to remote servers, especially in unstable or insecure environments which are common for IoT systems, may lead to unpredictable response latencies or confidentiality concerns~\cite{Sze2017,Shi2016}.
Moreover, transmitting a constant stream of data to the cloud is an energy-hungry operation, which can severely affect the battery life of the device~\cite{Zhou2019}.

For these reasons, \textit{extreme-edge} (i.e., on-device) computing has grown as an increasingly popular alternative for simple ML-based tasks~\cite{Shi2016,Zhou2019}.
Instead of a remote deployment on a high-end server, ML models are stored and executed directly on the device, eliminating or limiting the need to transmit the collected data.
This reduces both privacy and latency concerns tied to the unreliability of the Internet connection, while also possibly leading to higher energy efficiency.

Deploying ML at the edge is 
complicated by the tight resources budgets of IoT devices, which are mostly based on Microcontrollers (MCUs). 
Therefore, simple \textit{tree-based ensembles} such as Random Forests (RFs)~\cite{breiman2001} and Gradient-Boosted Trees (GBTs)~\cite{gbdt} are often regarded as a more lightweight alternative to state-of-the-art Deep Learning (DL) models in extreme-edge settings~\cite{gap8_ml}, since they can obtain comparable accuracy on simple tasks, with fewer parameters and operations per inference~\cite{shwartz2022tabular}.

Despite these advantages, the energy costs linked with tree ensembles inference can still be hard to sustain for battery-operated or energy-autonomous IoT nodes.
Accurate ensembles often include hundreds of Decision Trees (DTs), resulting in thousands of clock cycles per inference.
Several approaches have been introduced in the literature to optimize these models, generally consisting of pruning algorithms, which eliminate the least frequently used branches in each DT~\cite{ccp_pruning}.
However, these solutions modify the ensemble structure \textit{statically}, reducing its complexity once-for-all in exchange for a possible drop in accuracy. Thus, they offer limited flexibility in tuning the model execution costs at runtime.

In this work, which extends~\cite{adaptive_rf_conference}, we consider the much less explored path of \textit{runtime and input-dependent} optimizations for tree-based ensembles, motivated by the fact that: i) a system's energy budget may vary over time (e.g., depending on battery state), and ii) not all inputs require the same computational effort to achieve an accurate classification.
Indeed, most inputs are ``easy'', and a small subset of the DTs in the ensemble would be sufficient to classify them correctly, while saving energy.
On the other hand, statically shrinking the model would cause complex inputs to be wrongly labelled, negatively affecting the accuracy.

Accordingly, we study \textit{early stopping} policies that halt the execution of the ensemble after reaching a classification confidence target.
We use those policies to dynamically adapt the amount of computation to the system's requirements and to the difficulty of the processed data (stopping early for easy inputs), saving energy compared to a static ensemble.
While other works have studied dynamic inference for tree-based models~\cite{Gao2011,schwing2011,qwyc}, we are the first to thoroughly analyze the key issues and overheads associated with their deployment on a real-world, complex IoT platform. To this end, we design a tool that automatically generates optimized inference C code for both static and dynamic RFs and GBTs, starting from a Python model.
The following are our main contributions:
\begin{itemize}
    \item We introduce two novel early-stopping policies for dynamic inference of GBTs or RFs. Furthermore, we detail the deployment of these models on complex IoT devices, describing the required data structures and memory allocation techniques, while also exploring the effect of quantization on tree-based ensembles.
    \item  We study the effectiveness of early-stopping on \textit{multi-core} platforms, in which sets of DTs are evaluated in parallel, adapting our policies accordingly.
    \item We benchmark our dynamic models on three IoT relevant datasets, reducing an hardware-unaware estimate of time complexity from 57\% to 90\% with respect to static ensembles, with less than 1\% drop in accuracy on all the three tasks. When deployed on GAP8, a multi-core RISC-V architecture, our dynamic ensembles reduce the energy consumption by up to 42\% compared to a static RF/GBT, without losing accuracy.
\end{itemize}
The rest of the paper is structured as follows. 
Section~\ref{sec:background} provides the required background, and Section~\ref{sec:related} reviews the state-of-the-art; 
in Section~\ref{sec:methodology}, we present the details of the proposed early-stopping policies and of our implementation of dynamic tree-based ensembles for multi-core low-power platforms; lastly, Section~\ref{sec:results} reports the results of our experiments, and Section~\ref{sec:conclusions} concludes the paper.
\section{Background}\label{sec:background}
\subsection{Decision Trees}
Decision Trees (DTs) are shallow, non-parametric Machine Learning (ML) algorithms widely used for both classification and regression in supervised learning setups.
At training time (also known as ``growth''), these models learn a set of decision rules from the data, producing a piece-wise constant approximation of the target variable.
Specifically, starting from the root, each node compares one feature (column) of the input with a learned threshold and assigns the input either to its left or right child based on the result of such comparison. This process is repeated recursively until a terminal (leaf) node is reached, which contains the output estimate.
Since this work focuses on post-training and runtime optimizations of DTs, we omit a detailed description of the various fitting algorithms for DTs, referring readers to~\cite{dt_book} for further information.

\begin{figure}[ht]
    \centering%
    \includegraphics[width=0.7\linewidth]{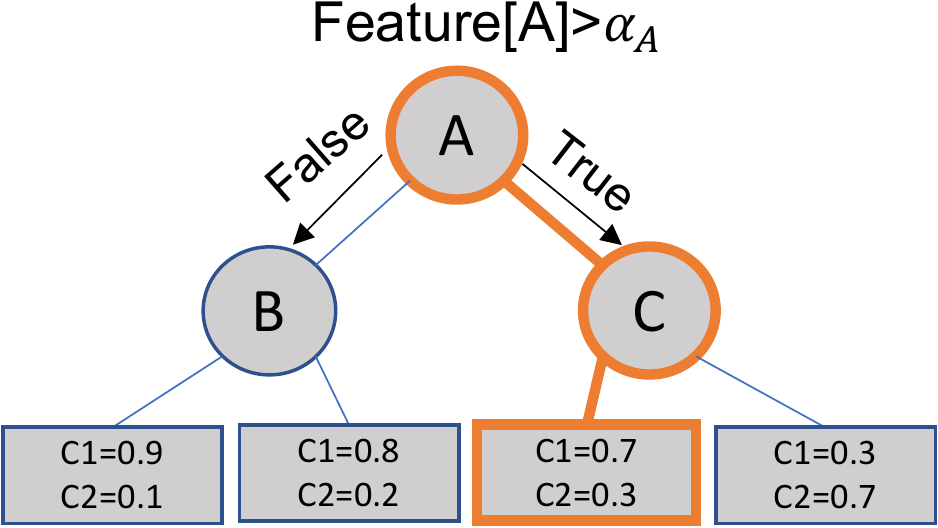}%
    \caption{Example of DT, where the root node performs a decision based on feature $A$ and threshold $\alpha_A$.}\label{fig:dt}%
\end{figure}

Figure~\ref{fig:dt} depicts a trained DT for a classification task, showing non-terminal nodes as circles and leaf nodes as rectangles.
Leaves can, in general, store either the class label or the entire array of class probabilities~\cite{gap8_ml}.
In case of regression, they contain the predicted scalar.

Algorithm~\ref{alg:dt_inference} reports the inference pseudo-code. 
We denote as Root($t$) and Leaves($t$), respectively, the root and the leaves of tree $t$. For each node $n$, Feature($n$) and $\alpha(n)$ are the input feature used for the split and its threshold,
while Right($n$) and Left($n$) are its descendants.
Lastly, Prediction($n$) extract the output value from the reached leaf.

\begin{algorithm}[ht]
\caption{Decision Tree Inference}\label{alg:dt_inference}
\begin{algorithmic}
\STATE 
\STATE $n=\mathrm{Root}(t)$
\STATE While $n \notin \mathrm{Leaves}(t)$
\STATE \hspace{0.5cm} if $\mathrm{Feature(n)} > \mathrm{\alpha}(n)$:
\STATE \hspace{1cm}  $n=\mathrm{Right}(n)$
\STATE \hspace{0.5cm} else:
\STATE \hspace{1cm}  $n=\mathrm{Left}(n)$
\STATE $P=\mathrm{Prediction}(n)$
\end{algorithmic}%
\end{algorithm}

The space complexity of a DT is $O(2^D)$, 
where $D$ is the \textit{depth}, i.e. the maximum-length path from the root to a leaf.
The upper bound is a \textit{perfect} tree with $2^D-1$ nodes.
The time complexity is $O(D + M)$, where $M$ denotes the number of classes (with $M=1$ for regression).
Reaching a leaf implies, at worst, $D$ branching operations, followed by an argmax over M elements to determine the largest output probability.

Due to their lightweight branching operations and limited memory requirements, DTs represent an ideal candidate for embedding inference on constrained edge nodes~\cite{stsensor}.
Nonetheless, these methods also have some shortcomings.
They are prone to overfitting
and tend to introduce a bias towards the majority class in unbalanced datasets~\cite{dt_book}.

\subsection{Tree-based Ensembles}\label{sec:back_ensemble}
In order to tackle these limitations, several DT ensembles have been introduced, in which multiple trees, referred to as ``weak learners'', perform an inference pass on the same input, before aggregating their output predictions.
This leads to sharp increases in accuracy and resistance to overfitting and unbalancing issues, at the cost of increased time and memory complexity.
We focus on the two most popular types of tree ensembles, i.e., RFs and GBTs.

\subsubsection{Random Forests}
RFs~\cite{breiman2001} are sets of classification DTs trained on a randomly selected subset of the data (i.e., with \textit{boosting}) and using only a random subset of the input features. This ensures diversity in the predictions and makes the RF less prone to overfitting. At inference time, each DT is applied to the input, and the output probabilities are accumulated. An argmax on the accumulated scores yields the final label.
Algorithm~\ref{alg:rf_inference} shows the corresponding pseudo-code, where TreeInference($t$) denotes a DT inference (i.e., Algorithm~\ref{alg:dt_inference}).

\begin{algorithm}[ht]
\caption{Random Forest Inference}\label{alg:rf_inference}
\begin{algorithmic}
\STATE 
\STATE $P =  \mathbf{0}_{M}$ // array of 0s of size $M$
\STATE for $t \in \mathrm{Forest}$:
\STATE \hspace{0.5cm}  $P = P + \mathrm{TreeInference}(t)$
\STATE $class =$ argmax(P)
\end{algorithmic}
\end{algorithm}

Noteworthy, for DT implementations that only store the predicted class label in leaf nodes, the RF aggregation can only use a crisp ``majority voting'', rather than a more precise averaging of probability scores. This is usually detrimental to accuracy; therefore, in this work, we follow the trend of most modern libraries~\cite{scikit-learn}, using weak learners that predict a probability value per class.

\subsubsection{Gradient-Boosted Trees}
the standard implementation of GBTs~\cite{gbdt} groups DTs in sets of cardinality $M$ called ``estimators'', conceptually executed in a sequence.
Each DT within an estimator is a \textit{regression} model, trained to predict the \textit{residual error} obtained by all previous estimators on a specific class.
At inference time, all DTs outputs are accumulated in a vector, which is then converted to probabilities with a formula that depends on the loss function used for fitting. As for RFs, the last step is an argmax to extract the label. Algorithm~\ref{alg:gbdt_inference} shows the pseudo-code of a GBT inference, where $t_i$ is the DT in charge of class $i$ within estimator $e$.
\begin{algorithm}[ht]
\caption{Gradient Boosting Trees Inference}\label{alg:gbdt_inference}
\begin{algorithmic}
\STATE 
\STATE $P =  \mathbf{0}_{M}$ // array of 0s of size $M$
\STATE for $e \in \mathrm{Estimators}$: // $e$ array of $M$ trees
\STATE \hspace{0.5cm} for $t_i \in \mathrm{e}$:
\STATE \hspace{1cm}  $P_i = P_i + \mathrm{TreeInference}(t_i)$
\STATE $class$ = argmax(compute\_probabilities($P$))
\end{algorithmic}
\end{algorithm}

\subsubsection{Complexity Analysis}
The space complexity of RFs and GBTs is $O(N*2^D)$ and $O(N*M*2^D)$, respectively, where $N$ is the number of estimators. For RFs, each single DT is considered an estimator, while for GBTs, an estimator is a set of $M$ trees, hence the additional multiplicative factor. Here, $D$ denotes the maximum depth across all DTs, which is generally fixed during training.
Similarly, the time complexity for inference, which is also linked with energy consumption, is $O(N*D)$ for RFs and $O(N*M*D)$ for GBTs.

\subsection{IoT End-node Target}
\label{subsec:gap8}
Microcontrollers (MCUs) are at the heart of most IoT end nodes, mainly due to their low production cost and high programmability.
In fact, while Application-Specific Integrated Circuits (ASICs) are potentially more energy efficient, especially for ML applications, their huge Non-Recurrent Engineering costs are unaffordable for most IoT solutions.

In recent years, the RISC-V Instruction Set has emerged in this domain due to its versatility and licensing-cost-free open-source nature~\cite{riscv_isa}.
In this work, we focus on the Parallel Ultra-Low-Power Processing Platform (PULP) family of RISC-V processors~\cite{gap8}, and specifically on the GAP8 System-on-Chip (SoC).
This SoC features one I/O core paired with an 8-core cluster, all leveraging an extended RISC-V instruction set with support for common signal processing and ML operations. 
The cores access a two-level memory hierarchy, including a 64 kB L1 with single-clock access latency (private of the cluster's cores) and a 512 kB L2. An additional L3 off-chip memory can be equipped to extend the storage capacity further but was not employed in this work.
GAP8 also features a general-purpose Direct Memory Access (DMA) controller to transfer data between memory levels, reducing access bottlenecks and allowing the programmer to control data transfers.

\subsection{Static and Dynamic ML Optimizations}

The problem of optimizing ML models to enable their execution on ultra-low-power edge nodes, trading off (small) accuracy drops for large latency, energy or memory savings, has been studied extensively in recent years, although with most focus being devoted to deep learning~\cite{anguita2012human, Daghero2021energy,daghero2021ultracompact, Jacob2018, JahierPagliari2018a, Tann2016}.

One broad characterization distinguishes \textit{static} and \textit{dynamic} optimizations. The former optimize a model before deployment, either during training or post-training. 
Among the most well-known static approaches are quantization and pruning~\cite{Daghero2021energy}, particularly popular for DL, which respectively limit the precision of data and operations or eliminate them, to improve both memory occupation and efficiency.

A fundamental limitation of static optimizations lies in their inability to adapt to \textit{changes in external conditions} during runtime, such as a low-battery state, or even more interestingly, to the processed input data.
Naively, this could be solved deploying multiple independent models (e.g., multiple RFs or multiple GBTs), each with a different trade-off in terms of accuracy vs energy/latency, and selecting among them at runtime. However, this approach would incur a large memory overhead, which is particularly critical for IoT end nodes.

Dynamic (or \textit{adaptive}) inference techniques, including this work, are designed to overcome these limitations. They allow the deployment of a single model able to adapt its complexity at runtime, while keeping the memory overhead under control~\cite{Tann2016, JahierPagliari2018a,har_journal}.
In practice, a dynamic model can be \textit{partially turned off} when the external conditions require it, or when the processing input's difficulty allows it~\cite{Daghero2021energy}. This partial shut-off can be realized in various ways, depending on the type of model considered~\cite{Tann2016, qwyc, Daghero2021energy}.
Most dynamic optimizations are \textit{orthogonal} to static ones, i.e., it is possible to build a dynamic system on top of statically optimized (e.g., quantized, pruned, etc.) ML models.

For dynamic ML systems that tune their complexity based on the input, a key component is a suitable \textit{policy}, i.e., the logic that selects which parts of the model to activate for a given datum~\cite{Park2015}. Good policies should be accurate but also incur low overheads. Section~\ref{sec:methodology} analyzes this aspect in detail.

\section{Related Works}\label{sec:related}
\subsection{Dynamic inference}
While dynamic/adaptive approaches are increasingly popular in the literature, the great majority applies solely to DL models. 
Most dynamic DL works adopt an \textit{iterative} approach, where the same input is processed multiple times, each time activating a larger ``portion'' of a neural network. After each iteration, the \textit{confidence} of the prediction is evaluated. The process is stopped when confidence reaches a pre-defined threshold.
This scheme assumes that easy inputs are the majority, thus most executions will stop at the initial iterations, reducing the average energy consumption.
On the other hand, complex inputs will still be classified by the largest ``version'' of the model, thus avoiding accuracy drops.
Literature works differ mainly in how they decompose the model. For instance, the authors of~\cite{Tann2016, har_journal, yu2018slimmable,JahierPagliari2018a} obtain a single sub-model by selectively deactivating a subset of the layers or channels of a network, or truncating the bit-width used to represent parameters.
Other works extend the approach more than two sub-models~\cite{Ngo2020a, Tann2016} or enhance the stopping criterion with class-aware thresholds~\cite{Daghero2021energy}.

Applications of adaptive inference to shallow ML classifiers are much less common.
In~\cite{schwing2011}, the authors propose an \textit{early stopping} criterion for tree-based ensembles, which models the prediction confidence after a binomial or multinomial distribution (depending on the number of classes), stopping the inference after a suitable subset of the trees has been executed. 
The authors benchmark their approach on seven small public datasets and a private one, showing a reduction of up to 63\% on the average number of trees executed with respect to the entire ensemble.
However, this approach requires the storage of large lookup tables in the order of $O(N^2)$, where $N$ is the number of estimators, thus incurring a significant overhead for large ensembles. %

In~\cite{Gao2011}, the authors leverage the partially aggregated probabilities of the already executed weak learners to determine the next tree to execute at runtime. 
This selection is performed according to multiple criteria: i) the current highest class probability and ii) the computational cost associated with each tree. 
Since weak learners within an ensemble process different features of the input datum, the inference cost is estimated taking into account not only the evaluation of the trees themselves, but also the extraction of any new feature that is not already available, i.e., that was not used by any of the previously executed weak learners.
A Gaussian distribution is used to obtain a probabilistic ``twin" of the classifier and determine when to trigger an early stop.
The authors also introduce a dimensionality reduction technique to limit the computations required to select the best next DT.
Nonetheless, the overhead of such a complex policy on an ultra-low-power device would be hard to sustain.
Indeed, as stated by the authors themselves, this approach becomes convenient only in the case of complex feature extraction, which is rarely the case in IoT applications~\cite{Gao2011}.

Lastly, the authors of~\cite{qwyc} propose the closest work to ours, introducing an early stopping method named Quit When You Can (QWYC). 
In this approach, two probability thresholds ($\epsilon_-$ and $\epsilon_+$) are extracted post-training, determining the boundaries to trigger an early stopping in binary classification tasks. At runtime, QWYC requires only two additional comparisons, introducing a minimal overhead.
Additionally, the authors propose a static sorting of weak learners, in which DTs able to trigger an early stopping most frequently are executed first.
However, QWYC is only evaluated on binary tasks, and no deployment results are provided.

\subsection{Tree-Based Ensembles Libraries}
Tree-based ensembles are widely used in various machine learning applications, and several optimized implementations have been proposed. Some works focus on optimizing inference time for high-end hardware~\cite{treelite, sklearn_porter, opencv_rf}, while others specifically target IoT edge nodes~\cite{gap8_ml,pulpissimo_rf}. In the former category, the authors of~\cite{opencv_rf} propose a C++-based implementation of RFs that supports both training and inference. They utilize an object-oriented representation of the trees, storing node information and thresholds ($\alpha$) in separate classes. However, they do not store class logits or support quantization, making their library less compact than those designed for IoT edge nodes. The implementation in~\cite{sklearn_porter} mirrors the DT data structures of~\cite{scikit-learn}, storing information such as child indexes, class logits, alpha values, and feature indexes for each node. Quantization is not supported in this case either. \cite{treelite} introduces a C++ implementation of RFs and GBTs. Single trees are implemented as classes, and nodes are represented as structures with pointers to left and right children, thresholds, and other fields. This implementation supports the integer representation of thresholds but only applied post-training and at 32-bit.
Despite being optimized for fast inference, these approaches are not suitable for IoT node deployment as they do not prioritize memory minimization, a crucial constraint for this type of device.

RF implementations tailored for RISC-V-based MCUs are presented in~\cite{gap8_ml,pulpissimo_rf}. The authors of \cite{pulpissimo_rf} benchmark various RF implementations on a single-core RISC-V MCU called PULPissimo, testing fully unrolled trees, recursive and for-loop-based inferences. Data storage is done using arrays or structures, and compiler-level optimizations are explored, resulting in up to 4$\times$ speed-up. In~\cite{gap8_ml}, the authors propose an array-based representation of trees, similar to our approach, specifically designed for the GAP8 SoC.

However, our work addresses several important aspects that have been overlooked in previous implementations. First, we store the logit values instead of just the predicted class, as they are necessary for enabling dynamic inference. Second, we discuss the allocation of the tree ensemble on a multi-level memory hierarchy. Finally, we enable various memory minimization techniques such as quantization at multiple precisions and optimized storage of children indexes, as described in Section \ref{sec:memory_alloc}. To the best of our knowledge, our library is the first to consider all these optimizations.
\section{Methodology}\label{sec:methodology}
\subsection{Early Stopping policies for Tree-Based Ensembles}\label{subsec:methodology_adaptive}

\subsubsection{Single-classifier Policies}\label{sec:single_scores}
So-called \textit{iterative} dynamic inference approaches~\cite{Daghero2021energy}, including ours, perform a sequence of classifications, either with different models or with different ``versions'' of the same model, deciding adaptively when to stop the process.
For these methods, most early-stopping policies use the output probabilities of the $t$-th classifier in the sequence ($P^t$) to determine the confidence of its prediction~\cite{Park2015, Tann2016, qwyc, har_journal}.

One of the most straightforward and computationally inexpensive approaches simply looks at the largest probability (i.e. the one associated with the most likely class).
Intuitively, a large top-probability will indicate a confident prediction and vice versa. We denote this policy as \textit{Max Score} ($s^t$).

While only requiring $O(M)$ comparisons per input, with M being the number of classes, this approach does not allow for a measure of the \textit{gap} between the top-probability and the others.
For instance, a 4-class output $P^t=[0.5, 0.5, 0, 0]$ corresponds to a large value for the metric ($s^t=0.5$), far from the random guess, but the classification is clearly highly uncertain, since $P^t_0 = P^t_1$.
In this case, using $s^t$ might mislead the early stopping into triggering too early, negatively affecting the accuracy.

A second policy that tries to overcome this issue is the \textit{Score Margin} ($sm^t$)\cite{Park2015, Tann2016}, which also considers the second largest probability in $P^t$ and is computed as follows:
\begin{equation}
    sm^t = \mathrm{max}(P^t) - \mathrm{max}_{\mathrm{2nd}}(P^t)
\end{equation}
While having the same $O(M)$ theoretical complexity, $sm^t$ requires approximately twice as many operations as $s^t$. On the other hand, it is generally more robust.
In the previous example, while $s^t=0.5$ may lead to wrong results, $sm^t=0$ clearly indicates that the classifier is not confident about its prediction, ensuring that the early stopping is not triggered.
Accordingly, $sm^t$ has become the most popular choice in recent literature~\cite{Park2015, Tann2016,har_journal}.

At runtime, $s^t$ or $sm^t$ are computed after each iteration, and compared with a user-defined threshold $t_h$. Using $sm^t$ as an example, the early stopping decision is formulated as:
\begin{equation}\label{eq:early_stop}
P =
\begin{cases}
    P^{0}\textrm{ if } sm^0 \ge t_h\\
    P^{1}\textrm{ if } sm^0 < t_h \wedge sm^1 \ge t_h\\
    P^{2}\textrm{ if } sm^0 < t_h \wedge sm^1 < t_h \wedge sm^2 \ge t_h\\
    ...\\ 
    P^{N-1}\textrm{ if } sm^{i} < t_h \textrm{, }\forall i < N
\end{cases}
\end{equation}
where $P$ is the final array of probabilities, which will be used to classify the input.
The energy versus accuracy trade-off is controlled by $t_h$, whose value alters the number of classifiers executed on average. Namely, a larger $t_h$ results in a more conservative system (giving higher priority to accuracy), and vice versa. Therefore, the threshold can be tuned at runtime to select different operating points based on external conditions, e.g., on battery state.

The main advantage of these confidence metrics is their low computational cost while also being accurate as long as the classifiers are well-calibrated~\cite{Guo2017}. Noteworthy, in case of a binary classification, $s^t$ and $sm^t$ become equally informative, since the second largest probability is just the complement of the largest.

\subsubsection{Aggregated Scores Policies}\label{sec:aggr_scores}
In their usual implementation, the metrics introduced in Section~\ref{sec:single_scores} are evaluated using only the probabilities produced by the \textit{last executed classifier} $t$, ignoring the outputs of previous models in the cascade~\cite{Park2015, Tann2016}.
This approach makes sense under the assumption that each new model is significantly more accurate than the previous ones, i.e., that $P^t$ is a much more reliable estimate of the true output probabilities with respect to $P^{t-1}$.

However, for ensemble models like RFs and GBTs, all weak learners (DTs) have comparable predictive power. It becomes then sub-optimal to decide for early stopping based only on the latest executed tree, ignoring the output of all previous ones. In light of this, we propose two extensions of the policies described in Section~\ref{sec:single_scores}, designed so that early stopping is triggered based on the \textit{accumulated predictions} of all weak learners already executed ($P^{[1:t]}$). In other words, we take a decision based on the aggregated prediction of the ``smaller ensemble'' composed of all already executed DTs.

The effectiveness of our approach lies in the fact that, for easy inputs, the accumulated probabilities quickly skew toward a single class after executing a small number of weak learners. Then, it becomes highly unlikely or even mathematically impossible for the leftover models to overturn the prediction, making their execution pointless to improve accuracy.

Mathematically, for an RF ensemble, we define the partial output after executing $t$ weak learners as:
\begin{equation}
P^{[1:t]} = \sum_{i=1}^{t}P^i
\end{equation}
We then define the Aggregated Max Score ($S^t$) policy as:
\begin{equation}
    S^t=\mathrm{max}(P^{[1:t]})
\end{equation}
and the Aggregated Score Margin ($SM^t$) as:
\begin{equation}
    SM^t= \mathrm{max}(P^{[1:t]}) - \mathrm{max}_{\mathrm{2nd}}(P^{[1:t]})
\end{equation}
The corresponding early stopping policies are obtained by replacing the array of probabilities of the last executed tree $P^t$ with the ones of \emph{all} executed trees $P^{[1:t]}$ and the score $sm^t$ with their aggregated versions $S^t$ or $SM^t$ in Eq.~\ref{eq:early_stop}.

For GBT, the formulation is similar except for one key difference. As mentioned in Section~\ref{sec:back_ensemble}, each estimator in a GBT is a set of \textit{regression} trees, whose outputs are converted to probabilities with a computationally expensive operation that depends on the training loss. Incurring the associated overheads after evaluating each estimator in order to extract $P^{[1:t]}$ could outweigh the benefits of early stopping. Thus, we leverage the fact that the conversion formula is \textit{monotonically increasing}\cite{scikit-learn}, and prefer to estimate confidence directly on the raw predictions.

Our results of Section~\ref{sec:results} show that the proposed aggregated scores policies obtain superior energy versus accuracy trade-offs with respect to state-of-the-art solutions that only account for the last learner.

\begin{figure}[ht]
    \centering
    \includegraphics[width=\linewidth]{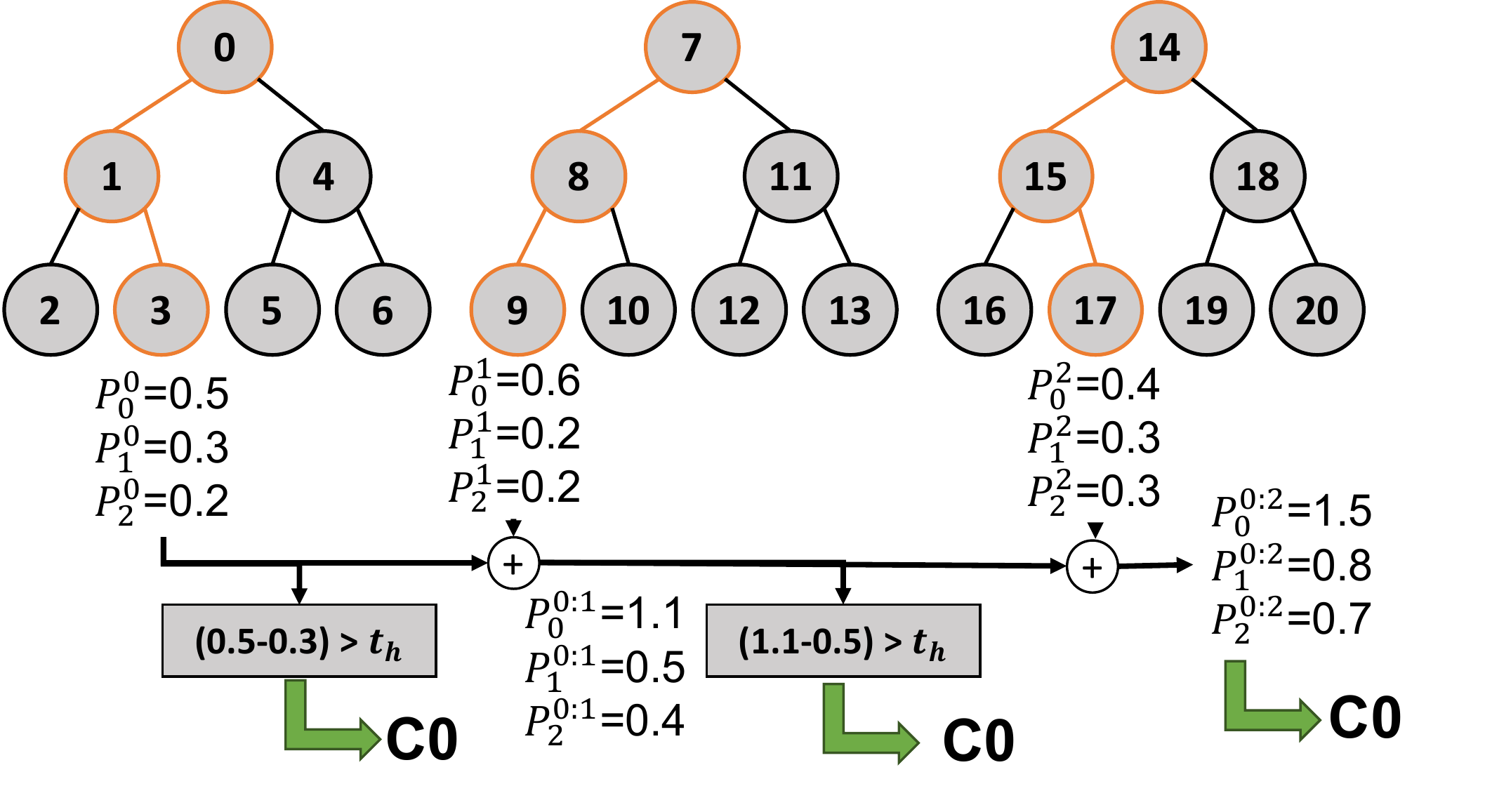}
    \caption{A dynamic RF with $N=3$, $M=3$ and $D=3$. In case early stopping is not triggered, the obtained output is identical to a static RF.}
    \label{fig:adaptive_rf}
\end{figure}

Figure~\ref{fig:adaptive_rf} shows a high-level overview of the adaptive inference mechanism proposed in this work, applied to an RF with $N=3$, $M=3$, $D=3$, and using $SM^t$ as confidence metric. We also assume a batch $B=1$ (more details on this in Section~\ref{subsec:methodology_adaptive_inference}).
Orange nodes represent the decision path taken in each tree for a hypothetical input.
After each weak learner, $SM^t$ is computed on the accumulated probabilities ($P^{[1:T]}$) and compared with the user-defined threshold $t_h$. As soon as $SM^t > t_h$, the process is stopped, and $P^{[1:t]}$ undergoes an argmax to extract the final predicted class $C_i$.

\subsection{Deploying tree-based ensembles on MCUs}
In this section, we describe our efficient library for static and dynamic RF/GBT inference on multi-core IoT end-nodes, such as our target GAP8, introduced in Section~\ref{sec:background}. 
Noteworthy, an RF library for GAP8 has recently been proposed in~\cite{gap8_ml}. However, its data structure is unsuitable for dynamic inference since it stores in the leaves only the most likely class rather than the full array of probabilities, making it impossible to derive confidence metrics. To our knowledge, there are no open-source GBT libraries for multi-core RISC-V MCUs.

For these reasons, we extend our previous in-house tool for the automated generation of optimized RF inference code~\cite{adaptive_rf_conference}, generalizing it to also support GBTs and to handle multi-core parallelism and complex memory hierarchy.
The tool outputs C code, generated with template programming starting from a Python model of the ensemble, and depending on its hyper-parameters ($N$, $M$, $D$, etc.)\footnote{The code is available open-source at: https://github.com/eml-eda/eden}.
The next sections describe the generated data structures (Section~\ref{sec:data_structures}), the memory allocation strategy (Section~\ref{sec:memory_alloc}) and the quantization employed to support our FPU-less target (Section~\ref{sec:quantization}). Note that while this work focuses on dynamic tree ensembles, our tool can also efficiently implement static models.
\subsubsection{Ensembles structure}\label{sec:data_structures}
\begin{figure}[ht]
    \centering
    \includegraphics[width=\linewidth]{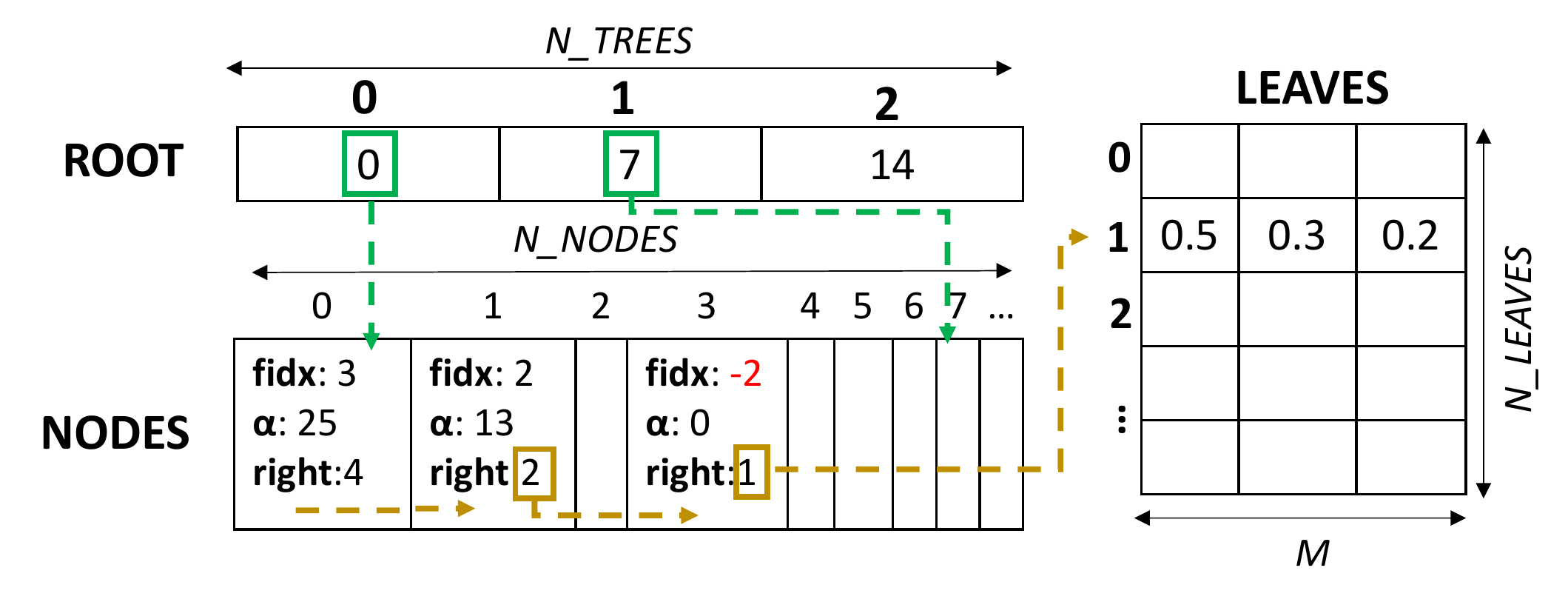}
    \caption{C data structures of our tree ensemble library in the case of a RF. The arrows represent the inference steps for the first tree in Figure~\ref{fig:adaptive_rf}.}
    \label{fig:rf_struct}
\end{figure}

Our data structures take inspiration from the open-source OpenCV~\cite{opencv_rf} library, with several modifications to make them more efficient for low power MCUs.
Specifically, we replace lists with C arrays, saving memory and improving data locality while also making the structure more compact.
Figure~\ref{fig:rf_struct} shows the three main structures for a RF with $M=3$ classes.
The NODES array is composed of C ``structs'', representing the information of all DT nodes.
Each node has three fields:
\begin{itemize}
    \item $fidx$: storing the index of the input feature considered by the node. At inference time, it is used to select the input value compared with the threshold $\alpha$ to determine the next visited node. For leaves, this field is set to the special value -2 for compatibility with~\cite{scikit-learn}.
    \item $\alpha$: the threshold compared against the input value at position $fidx$. If the latter is smaller or equal (larger) than $\alpha$, we visit the left (right) child next.
    \item $right$: the offset in NODES between the current node and its right child. For terminal nodes, we reuse this field to store a row index in the LEAVES matrix, holding the class probabilities assigned to samples reaching that leaf.
\end{itemize}
The ROOTS array stores the indexes of the root nodes of each tree in NODES, allowing a fast iteration among the trees. Lastly, as mentioned, the LEAVES matrix stores the class probabilities of all leaves.

The inference pseudo-code for a single tree, in the most general case of a multi-class RF, is shown in the ``run\_tree'' function of Algorithm~\ref{alg:dt_inference_deployed}. Note that we do not store the index of the left child of a node, to save memory.
Instead, we organize our data structure so that the left child for all non-leaf nodes is always (implicitly) the next element in the NODES array. This is obtained by generating the structure during a \textit{pre-order} visit of each tree.
The special value in $fidx$ indicates when a leaf has been reached, thus being used as a loop exit condition.
C denotes the total number of cores available during the inference, which will be discussed in detail in Section~\ref{sec:multicore}.

\begin{algorithm}\caption{Static multi-class RF inference pseudo-code}\label{alg:dt_inference_deployed}
\begin{lstlisting}[language=C++]
|\textcolor{blue}{run\_tree}|(t, P, INPUT, ROOTS, NODES, LEAVES) {
  if (core_id == (t%
    n=NODES[ROOTS[t]];
    while(n.fidx != -2) {
      if(INPUT[n.fidx]>n.alpha) n+=n.right;
      else n=n+1;
    }
    |\textcolor{magenta}{critical\_section\_in}|();
    for(j=0;J<M;j++) P[j]=P[j]+LEAVES[n.right][j];
    |\textcolor{magenta}{critical\_section\_out}|();
  }
}

// L1 Memory -> INPUT, P, ROOTS
// L1 or L2 Memory -> NODES,LEAVES
P = {0};
parallel for (t=0; t<N; t++)
    |\textcolor{blue}{run\_tree}|(t, P, INPUT, ROOTS, NODES, LEAVES);
|\textcolor{magenta}{barrier}|();
if(core_id == 0) res = |\textcolor{blue}{argmax}|(P);
\end{lstlisting}
\end{algorithm}

We further optimize our data structures when working with \textit{binary} RF classifiers or GBTs. In the first case, each leaf needs only to store a single class probability (since $P_1 = 1-P_0$). Thus, we can save this value directly in the $\alpha$ field of the leaf, completely removing the LEAVES array.
Similarly, GBTs regression trees require the storage of a single value per leaf, allowing us to apply the same optimization.

\subsubsection{Memory Allocation Strategy}\label{sec:memory_alloc}
Modern IoT end nodes, including our target, have complex multi-level memory hierarchies.
In particular, many of these devices use software-controlled scratchpad memories rather than hardware caches, coupled with Direct Memory Access (DMA) controllers to move data between, for instance, a smaller but faster L1 memory, and a bigger but slower L2 memory. With respect to using hardware caches, this approach requires more effort on the software side, but results in smaller and more power-efficient hardware, which is crucial for IoT nodes, while also possibly providing performance benefits for applications characterized by predictable and regular memory access patterns, such as many ML models. Examples of these devices are found both in academia~\cite{diana} and in commercial products~\cite{gap8,gap9}.

Maximizing L1 accesses is, therefore, imperative to reduce inference latency and energy.
The problem is not trivial, since ensembles achieving high accuracy, even for relatively simple tasks such as those considered in Section~\ref{sec:results}, are generally too large to fit entirely in L1 (GAP8, for instance, has a 64kB L1).

One solution would be to employ a \textit{tiling} approach, dynamically loading to L1 only the data required to execute a small chunk of computation (e.g., a single tree inference). This is the approach generally taken by DL libraries for edge devices~\cite{dory}. The regularity of neural network computations makes tiling a profitable option because: i) data portions needed in L1 can be statically determined at compile time, and ii) once loaded, \textit{all} data elements will be accessed and reused multiple times, amortizing the transfer overheads.

On the contrary, for tree-based ensembles, the access ratio of the NODES structure is logarithmic, requiring the transfer of up to $2^D$ nodes per tree, but accessing at most $D$ elements, with at most one access per node. Thus, the data transfer overhead out-weights the benefits of having node information in L1, making tiling detrimental. Similar considerations apply to the LEAVES matrix, whose rows are accessed with an increasing yet randomly strided and sparse pattern (1 every $2^D$ rows in the worst case). In contrast, the input sample array (INPUT in Algorithm~\ref{alg:dt_inference_deployed}) is reused by all DTs in the ensemble, and multiple nodes within each tree might access the same element. Similarly, the array of accumulated outputs (P in Algorithm~\ref{alg:dt_inference_deployed}) is accessed densely and with a regular pattern at the end of each DT inference.

We define a static (compile-time) memory allocation strategy for our tree ensemble code generator based on these considerations. We load INPUT, P, and the ROOTS array (whose size is generally negligible, i.e., less than 1kB) entirely in L1. We then compute the leftover L1 memory and check if the LEAVES or NODES structures can fit in the remaining space, prioritizing the former. When this happens (for small ensembles), all required structures are stored in L1. Otherwise, NODES and LEAVES are directly accessed from L2. Lines 14 and 15 of Algorithm~\ref{alg:dt_inference_deployed} summarize the allocation scheme. We verified experimentally that this produces a faster and more efficient inference than tree-wise tiling.

Note that this proposed memory allocation strategy is valid for any device characterized by a multi-level memory and a software-managed caching mechanism. Changing the deployment target only impacts the dimension of L1, which has to be specified as an input argument for our allocation strategy. On the contrary, SoCs equipped with hardware-controlled caches can skip this memory placement step.

\subsubsection{Data Quantization}\label{sec:quantization}
One of the most promising approaches to make ML models compatible with edge devices is quantization, an optimization which consists of reducing the precision used to store inputs and parameters~\cite{Jacob2018}.
This reduces memory occupation and improves speed and energy efficiency for IoT end-nodes, where FPUs are either slower and more energy-hungry than ALUs, or completely absent, as in the case of GAP8, causing floating point operations to be approximated with expensive software routines.

While extensively studied for DL~\cite{Jacob2018}, quantization is much less explored for tree ensembles. For RF/GBT classifiers, the valid targets for quantization are: i) the input array, ii) the internal comparison thresholds of each DT node ($\alpha$), and iii) the output probabilities. Since i) and ii) are directly compared, they should be quantized with the same precision and format.

Input and threshold quantization can be introduced at training time (so-called \textit{quantization-aware training}) by simply converting inputs to integers before starting the process. The comparison thresholds generated by the training framework~\cite{scikit-learn} will still be floats in general. However, given that inputs are integers, it can be easily seen that if the thresholds are quantized by simply truncating their fractional part, the nodes' decisions will not be altered.

In contrast, our tool quantizes the leaves probabilities after training (a.k.a., \textit{post-training quantization}), statically computing the range of the values that the accumulated probabilities can assume, and using it to determine the quantizer parameters.

In both cases, we use a symmetric min-max quantizer~\cite{Daghero2021energy}, computed with the following equations:
\begin{align}
    x_{int} &= round \left(\frac{x \cdot 2^{bits-1}}{\max(|x|)} \right) \\
    x_{Q} &= clamp(-2^{bits-1}, 2^{bits-1} -1, x_{int}) \\
\end{align}
Where $x$ is the floating point value, and the max is computed over all training samples. The $clamp$ is necessary for outliers that fall outside the training range and is defined as follows:
\begin{equation}
    clamp(a,b,x) =  \begin{cases}
a &\text{if $x \leq a$}\\
x &\text{if $a \leq x \leq b$}\\
b &\text{if $x \geq b$}
\end{cases}
\end{equation}

We find that the accuracy loss when quantizing inputs, thresholds and outputs is often negligible. The detailed trade-off between quantization bit-width (8, 16, 32 bits for inputs, thresholds, and leaves) and accuracy is analyzed in Section~\ref{subsubsec:results-quantization}.

\subsection{Multi-core inference}\label{sec:multicore}

\subsubsection{Static ensembles}
\begin{figure}[ht]
    \centering
    \includegraphics[width=\linewidth]{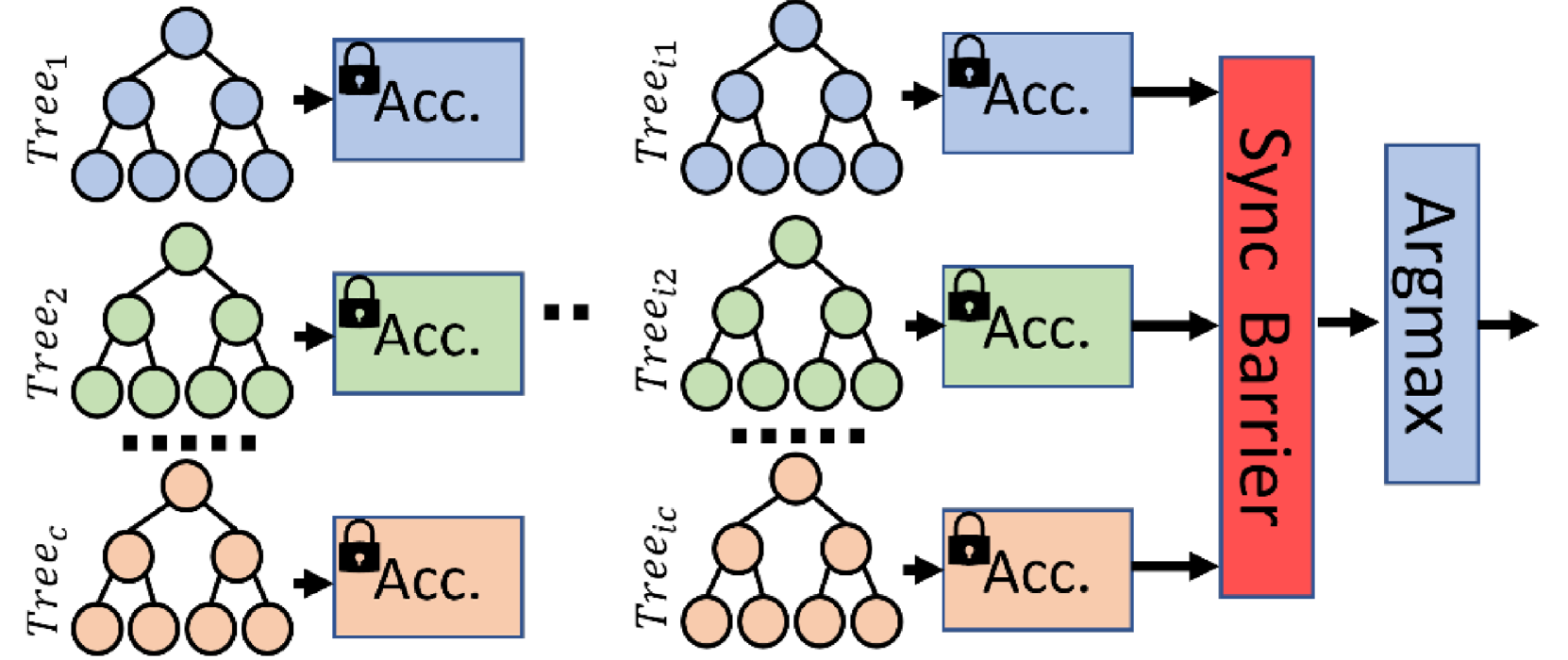}
    \caption{Multi-core inference for a static tree ensemble.}
    \label{fig:methodology_multicore_ensemble_static}
\end{figure}

To parallelize static RFs/GBTs on multi-core IoT platforms, we use the approach proposed in~\cite{gap8_ml} as a starting point.
Figure~\ref{fig:methodology_multicore_ensemble_static} schematizes a static inference on $C$ cores (each represented by a different color), which corresponds to the pseudo-code of Algorithm~\ref{alg:dt_inference_deployed}.
DTs are statically assigned to a core based on their index in the ensemble. 
Mutual exclusive access, indicated by a lock (\texttt{critical\_section} in Alg.~\ref{alg:dt_inference_deployed}), is required when accumulating probabilities on the shared output vector P (\textit{Acc.} in the Figure). Finally, a barrier has to be inserted after the parallel execution of trees, before the final argmax computation, performed only by Core0.

Noteworthy, this scheme does not enforce a specific order on the DT executions in different cores. Global synchronization is required only at the end. In case of GBTs, also DTs belonging to different estimators can run in parallel.

\looseness=-1
\subsubsection{Dynamic ensembles}\label{subsec:methodology_adaptive_inference}
Previous dynamic inference approaches for tree ensembles~\cite{Gao2011,qwyc,schwing2011} evaluate the early-stopping policy (Section~\ref{subsec:methodology_adaptive}) after executing each DT. However, as shown in our previous work~\cite{adaptive_rf_conference}, this is not necessarily optimal. Evaluating the policy more rarely (thus reducing the associated overheads due to its computation, i.e., Equation~\ref{eq:early_stop}) might give benefits superior to the occasional wasted energy for executing ``useless'' extra DTs.

This becomes even more relevant in multi-core setups, where performing a stopping decision after each DT is highly sub-optimal. In fact, $C$ trees are concurrently being executed at all times (with $C=$ number of cores), requiring, on average, a similar execution time. 
Thus, taking an 8-core system for example, halting after either 10 or 16 DTs consumes almost the same amount of time and energy. However, the first option may result in a less informed decision, as it disregards the output of the remaining 6 trees, which is likely already available or produced shortly. 
Noteworthy, these considerations are ignored by all previous works, which assume a purely sequential computation model~\cite{schwing2011, Gao2011,qwyc}.

\begin{figure}[ht]
    \centering
    \includegraphics[width=\linewidth]{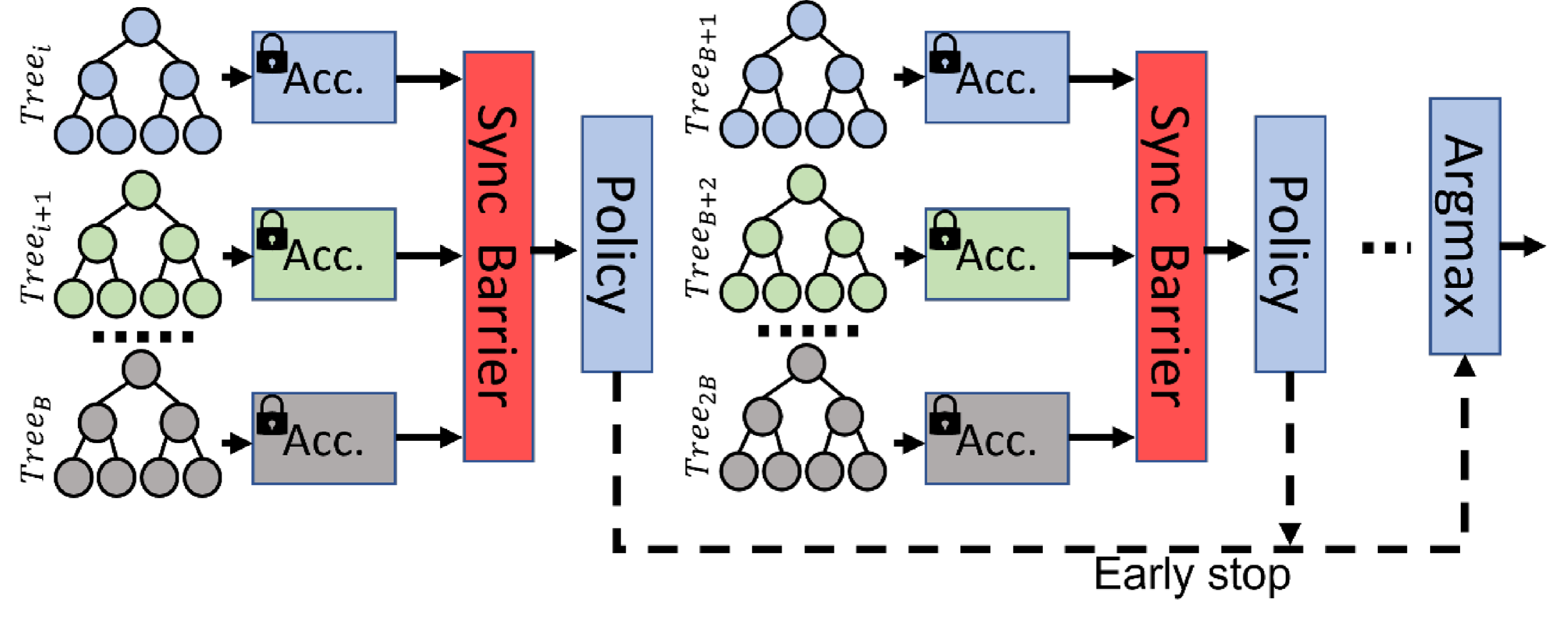}
    \caption{Multi-core inference for a dynamic tree ensemble.}
    \label{fig:methodology_multicore_ensemble_dynamic}
\end{figure}

In contrast, we follow these observations and propose a configurable \textit{batching} mechanism, in which early-stopping is considered only after all cores have executed their following estimator. Figure~\ref{fig:methodology_multicore_ensemble_dynamic} and Algorithm~\ref{alg:adaptive_inference_deployed} schematize this approach for an RF. 
In the pseudo-code of Algorithm~\ref{alg:adaptive_inference_deployed}, each iteration of the outer loop in lines 4-11 corresponds to one of these macro-steps, whose maximum number is computed statically and inserted in the POLICY\_TRIGGERS constant. Lines 12-15 handle the final ``left-over'' DTs when the total $N$ is not a multiple of the batch size. The ``policy'' function in line 9 represents the evaluation of $S^t$ or $SM^t$, whose value is compared with $t_h$ to set the stop flag.

Compared to the execution of static ensembles, an additional barrier is inserted after each batch of $B$ trees, allowing the execution of the early-stopping policy, which is in charge of Core0. When the policy determines that inference should be halted, execution jumps directly to the final argmax. 
Noteworthy, the added barriers (lines 7, 10), the computation of the policy, and its comparison with the exist threshold (line 9), cause an overhead in terms of latency and energy in the dynamic ensemble, which however is often minimal, as detailed in Sec.~\ref{sec:results}.

\begin{algorithm}\caption{Dynamic multi-class RF inference pseudo-code}\label{alg:adaptive_inference_deployed}
\begin{lstlisting}[language=C++]
P = {0};
t = 0;
stop = 0;
for(int bt=0; bt < POLICY_TRIGGERS && !stop; bt++) {
  for (int i = 0; i < B; i++)
    |\textcolor{blue}{run\_tree}|(t++, P, INPUT, ROOTS, NODES, LEAVES);
  |\textcolor{magenta}{barrier}|();
  if (core_id == 0)
    stop = |\textcolor{blue}{policy}|(P) > th;
  |\textcolor{magenta}{barrier}|();
}
if (!stop) {
  while(t<N)
    |\textcolor{blue}{run\_tree}|(t++, P, INPUT, ROOTS, NODES, LEAVES);
}
|\textcolor{magenta}{barrier}|();
if (core_id == 0) res = |\textcolor{blue}{argmax}|(P);
\end{lstlisting}
\end{algorithm}

We set the batch size $B$ equal to the available cores $C$ to ensure that all hardware resources are fully used. In the case of RFs, where an estimator corresponds to a single DT (the total number of trees is identical to N, i.e., the number of estimators), we perform an early stopping decision once every B executed trees. For GBTs, instead, early stopping decisions can only be performed after executing an entire estimator, i.e., a group of $M$ trees, (the total number of trees in the ensemble is $N\cdot M$, with $N$ being the number of estimators, and $M$ the number of classes, see Section~\ref{sec:background}).

\section{Results}\label{sec:results}
\subsection{Target Benchmarks}
We benchmark our work on three diverse IoT-relevant tasks: surface Electromyography (sEMG)-based hand gestures recognition, hard-drive failure detection and Human Activity Recognition (HAR) based on accelerometer data.

For sEMG-based gesture recognition, we employ the \textbf{Ninapro DB1}~\cite{ninapro}, which encompasses EMG signals collected from 27 healthy subjects while performing hand movements.
We follow the experimental setup described in~\cite{ninapro}, using the same pre-processing and data split, considering 14 hand movements classes, and a 10-channel EMG signal as input.
We use a window of 150 ms, collected at 100 Hz, thus obtaining a dataset with $\approx$207k elements.
As in most state-of-the-art works~\cite{ninapro}, we use a patient-specific training procedure, i.e., we train separate models for each subject in the dataset, using different recording sessions as training, validation and test sets.
For sake of space, we show graphical results only for the first two subjects (S1 and S2), while reporting aggregate metrics over all 27 subject in tables.

For hard-drive failure detection, we analyze the \textbf{Backblaze}~\cite{BackblazeHardDrive} dataset, containing 19 Self-Monitoring Analysis and Reporting Technology (SMART) features collected from hard disks by different vendors during their lifetime in a data center from 2014 to 2019. 
The goal is predicting whether a disk will experience a failure in the next 7 days.
For this dataset, we mirror the setup shown in~\cite{burrelloPredictingHardDisk2021} in terms of data split, preprocessing, and feature selection.
Namely, we feed models with 90-day windows of the 19 features (each feature has 1 sample per day), obtaining a dataset with $\approx$707k elements.
We use 10\% of the training data as validation set.

Lastly, we consider the \textbf{UniMiB-SHAR}~\cite{micucci2017unimib} HAR dataset, featuring 3-axis acceleration signals collected from smartphone accelerometers, during 9 different daily-life activities (e.g., walking, standing, etc.) and 8 different kinds of falls.
The sampling frequency is 50 Hz, and the authors provide the data already pre-processed in windows of 151 samples ($\approx$3s) centered around acceleration peaks. The datasets contains around 11k elements.
We benchmark our models on the AF-17 task~\cite{micucci2017unimib}, which considers all 17 classes without subject-specific training, using the default pre-processing and windowing.
Samples are divided into training, validation, and test datasets with a 60\%, 20\%, 20\% split.

The tasks involve different kinds of inputs signals, input dimensions (from 150 ms in NinaPro to 90 days in BackBlaze), and number of classes (from 2 in BackBlaze to 17 in UniMiB-SHAR), leading to RF/GBT models whose complexity spans over 3 orders of magnitude. 
Due to the unbalanced nature of the training sets, we augment the training sets by performing an oversampling of the minority classes.

In the following sections, we report our results using the top-1 macro average accuracy (also known as balanced accuracy, i.e., the average of each class recall) for Ninapro and UniMiB-SHAR and the F1-score for Backblaze.

\subsection{Experimental Setup}\label{subsec:results-setup}

All ensembles have been trained using Python 3.8 and the Scikit-Learn~\cite{scikit-learn} library. 
To build our comparison baseline, we explore with grid search all static RFs and GBTs with the following combinations of hyper-parameters: depths in the range [1,15], number of estimators in [1,40], and input and leaves quantization to 8/16/32 bits, for a total of 5400 architectures tested for each dataset and model type.
For Ninapro, given the personalized training, we repeated the grid search for each of the 27 subjects.
For RFs on Backblaze, we instead fixed the maximum depth of the ensembles to 38 and limited the number of estimators to less than 30, following the reference work of~\cite{burrelloPredictingHardDisk2021}.
After each search, we excluded static models too large to fit the L2 
 512kB memory of GAP8, and selected the top scoring one on the validation set as starting point to derive our dynamic model.
Section \ref{sec:static_inference} reports the results of this grid search, in which we estimate time complexity using a hardware-agnostic metric, i.e., the average number of visited tree nodes per inference.

Sections \ref{sec:dynamic_hw_unaware}-\ref{sec:deployment} analyze dynamic solutions: in Sec. \ref{sec:dynamic_hw_unaware}, we report hardware-agnostic results with all dynamic policies; in Sec. \ref{sec:ordering}, we discuss the impact of execution order on dynamic ensembles; lastly, in Section \ref{sec:deployment}, we report the results obtained deploying all the dynamic and static models that are Pareto optimal in terms of scoring metric (Accuracy or F1) versus memory or time complexity.
All deployments use our automated code generation tool, and target the GAP8~\cite{gap8} SoC introduced in Section \ref{subsec:gap8}. We set both the cluster and the fabric controller clock frequencies to 100 MHz.
The inference runs entirely on the cluster cores.

\subsection{Static Inference Results}
\label{sec:static_inference}
\begin{figure*}[t]
\centering
\includegraphics[width=0.9\linewidth]{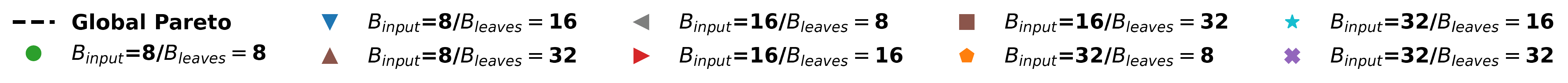}
\includegraphics[width=0.85\linewidth]{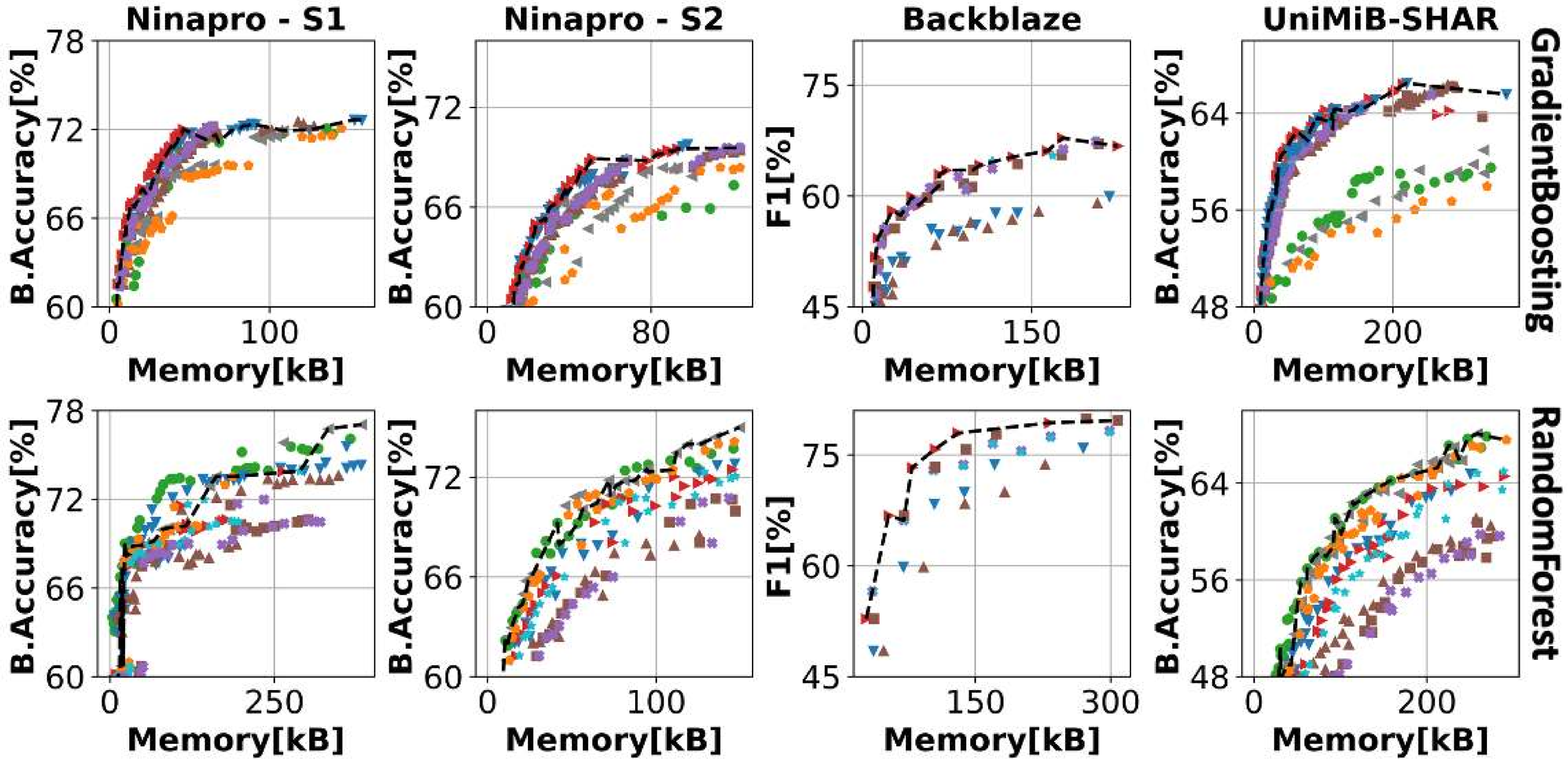}
\caption{Pareto front of score vs. memory occupation for ensembles with quantized inputs ($B_{input}$) and outputs ($B_{leaves}$) on the validation set scored on the test set.}
\label{fig:results-quantization}
\end{figure*}
In this section, we report the results of the grid search for static RFs and GBTs on the three target tasks, with the goal of analyzing the trade-offs among the two types of models.
\subsubsection{Ensembles quantization}
\label{subsubsec:results-quantization}
Figure~\ref{fig:results-quantization} shows the models on the score vs. memory occupation Pareto front extracted from the validation set at different bit-widths for inputs/thresholds ($B_{input}$) and outputs ($B_{leaves}$) and scored on the test sets.
For all datasets, we notice that points obtained with 8-bit output quantization are never on the global Pareto front for GBT, with Backblaze models incurring a F1 drop so large that they are omitted from the figure for easier visualization. This is probably due to the wider ranges of the GBT outputs. On the contrary, 8/16-bit inputs and 16-bits outputs are generally achieving the best memory versus score trade-offs. 
Concerning RFs, fewer bits are generally required, since leaf nodes store probabilities, with narrower ranges.
In this case, 8-bit quantization is often enough, both for inputs and outputs.
The only exception is represented by Backblaze, where 8-bit quantization causes sharp decreases in F1 score. 
For both types of models, we observe that 32-bit ensembles are rarely on the Pareto fronts. We impute this behavior to the combination of: i) the regularization effect of quantization, which, as already observed in Neural Networks~\cite{Jacob2018}, can lead to better generalization, and ii) the significant increase in memory that 32-bit models incur, leading rapidly to exceeding the L2 of the target device.
\subsubsection{RF vs GBT comparison}
Fig. \ref{fig:results-quantization} also shows the global static Pareto fronts in the scoring metric versus memory occupation space.
Specifically, we extract the Pareto points from the validation set, reporting then their score on the test set.
On all datasets, we observe that for lower memory footprints (less than 40/150 kB, depending on the task), GBTs tend to outperform RFs, achieving higher accuracy for the same space occupation.
Vice versa, RFs outperform GBTs under less tight constraints, while also reaching the highest score values for models fitting GAP8's memory on all tasks.
On the Ninapro DB1 dataset, for S1, RFs reach up to 77.05\% of balanced accuracy (vs 72.64\% of GBTs), while for S2, they achieve 74.98\% of accuracy (vs 69.56\%). On Backblaze,
RFs achieve a maximum F1 score of 79\%, compared to the 66\% achieved by the best GBT model;
Lastly, for UniMiB-SHAR, RFs obtain a 2\% higher maximum accuracy (67\% vs 65\%), but GBTs perform significantly better in the low-memory regime (e.g., the smallest GBT reaching 52\% requires 4x less memory than the smallest RF achieving the same score).
This trend is a direct effect of the structure of the two model types; GBTs do not need an external leaves array to store the probability of all the output classes, as discussed in Section~\ref{sec:methodology}, thus requiring less memory. This saving is more evident for smaller models, in which the LEAVES array size is comparable to the one of the NODES structure.

Fig.~\ref{fig:results-adaptive-branches} shows the trade-off between the scores achieved by the models and the \textit{number of visited nodes} per inference, averaged on all input samples. We use this metric as an estimate of the time and energy complexity for an inference, more accurate than just counting the number of DTs, since our models also have varying depths. The Pareto optimal models shown in this figure are in general distinct from those in Figure~\ref{fig:results-quantization}. In contrast to memory occupation, static RFs always outperform static GBTs in terms of time complexity, achieving gains ranging from 2$\times$ to 45$\times$ at iso-accuracy for Ninapro, and from 4$\times$ to 30$\times$ for UniMiB-SHAR. This is because, each GBT estimator includes one regression tree per class (vs. a single classification tree for RFs).
Only on hard disk failure detection, which is indeed the task with the smallest number of classes (two), the trend is similar to the memory one.

Overall, these results show that while GBTs are generally outperformed in terms of inference time complexity, they are competitive for small memory budgets.
For this reason, we explore dynamic inference for both types of ensembles.

\subsection{Dynamic Inference: Hardware-agnostic Results}
\label{sec:dynamic_hw_unaware}
\begin{table*}[ht]
\centering
\begin{tabular}{l|l|l|l|l|l|l|l}
\textbf{Dataset}                   & \textbf{Depth}                & \textbf{\#VisitedNodes}      & \textbf{\#Estimators} & \textbf{$B_{input}$}   & \textbf{$B_{leaves}$}  & \textbf{Score [\%]}       & \textbf{Memory [kB]}    \\ \hline\hline
\multicolumn{1}{c}{\textbf{GBTs}} & \multicolumn{1}{c}{\textbf{}} & \multicolumn{1}{c}{\textbf{}} & \multicolumn{1}{c}{}  & \multicolumn{1}{c}{} & \multicolumn{1}{c}{} & \multicolumn{1}{c}{} & \multicolumn{1}{c}{}  \\ \hline
Ninapro                            & 5.9   [$\pm$0.7]                   & 3060   [$\pm$387]                  & 37[$\pm$3.5]               & 17[$\pm$10]               & 20[$\pm$9]                & 75[$\pm$6]           & 199.5[$\pm$65]                 \\
Backblaze                               & 15                            & 128.53                            & 9                     & 16                   & 16                    & 66                  & 226                           \\
UniMiB-SHAR                        & 8                             & 2987                          & 22                    & 8                    & 16                   & 65                 & 363                             \\ \hline
\multicolumn{1}{c}{\textbf{RFs}}   & \multicolumn{1}{c}{\textbf{}} & \multicolumn{1}{c}{\textbf{}} & \multicolumn{1}{c}{}  & \multicolumn{1}{c}{} & \multicolumn{1}{c}{} & \multicolumn{1}{c}{} & \multicolumn{1}{c}{} \\ \hline
Ninapro                            & 13.8   [$\pm$1.28]                 & 348   [$\pm$81]                    & 31.8   [$\pm$7]         & 16   [$\pm$9.5]         & 12.4   [$\pm$4.5]         & 77   [$\pm$6]        & 335.83   [$\pm$97]       \\
Backblaze                               & 38                            & 155                            & 9                     & 16                   & 32                    & 79                 & 308                         \\
UniMiB-SHAR                        & 15                            & 136                           & 10                    & 32                   & 8                    & 67                 & 292         \\ \hline               
\end{tabular}
\caption{Static RFs/GBTs used as a starting point to construct dynamic ensembles.}\label{tab:adaptive_models}
\end{table*}
\begin{figure*}[ht]
\centering
\includegraphics[width=0.8\linewidth]{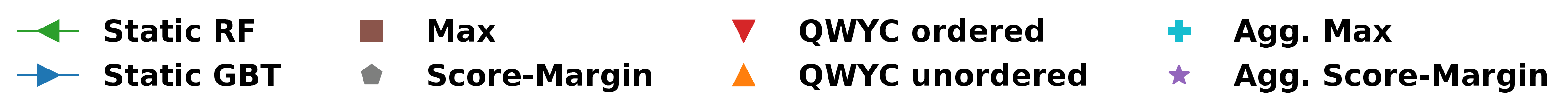}
\includegraphics[width=0.8\linewidth]{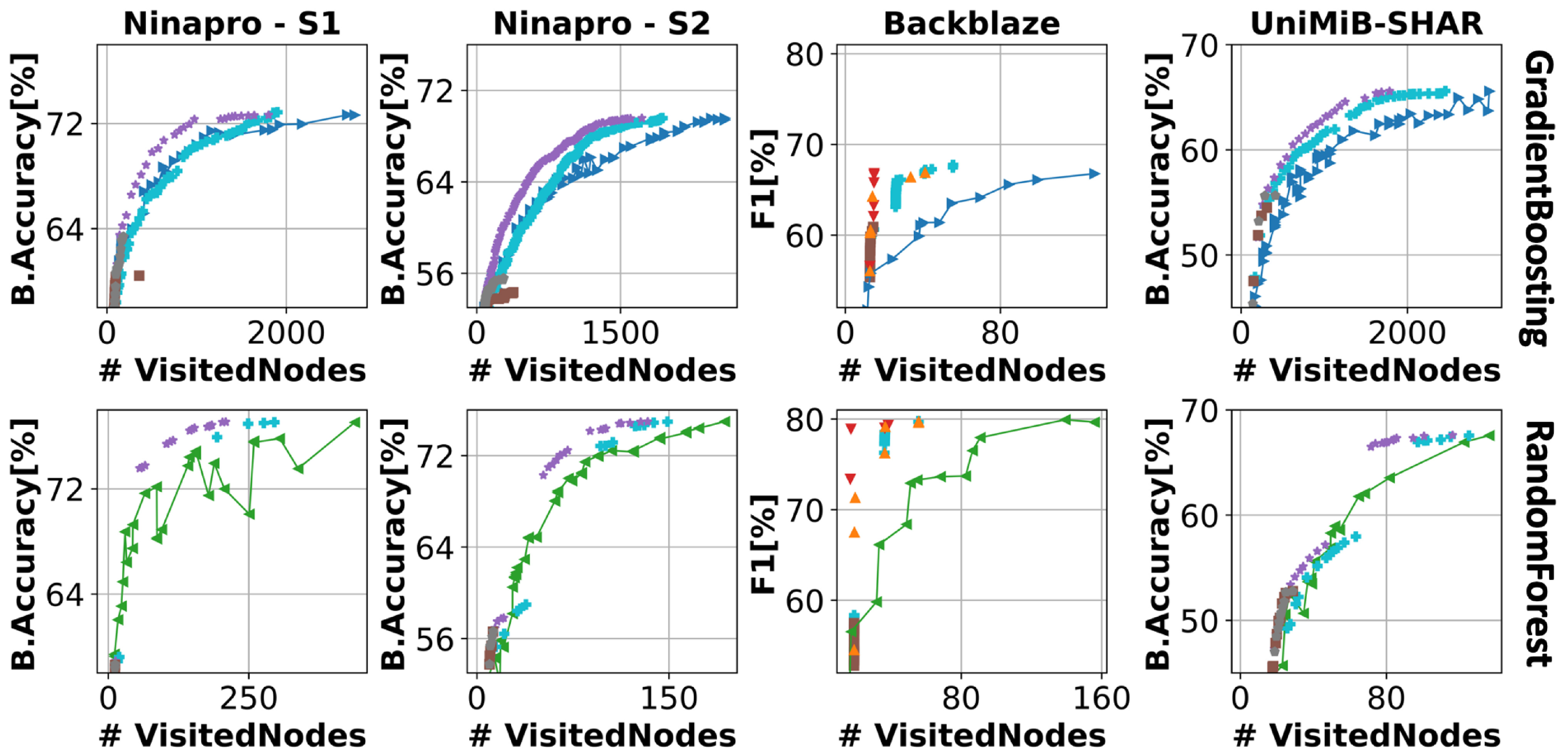}
\caption{Dynamic and static ensembles Pareto fronts obtained on the validation set and scored on the test set with batch $B=1$.}
\label{fig:results-adaptive-branches}
\end{figure*}

In this section, we discuss the results obtained with our proposed dynamic inference policies (Agg. Max and Agg. Score-Margin), comparing them against the static models discussed in the previous section and against three state-of-the-art dynamic policies, namely Max, Score-Margin, and Quit-When-You-Can (QWYC). 
Specifically, the comparison is done following the setup described in Section~\ref{subsec:results-setup}, and reported in terms of accuracy versus the average number of visited nodes as a proxy for time/energy complexity, since the goal of early-stopping adaptive models is precisely to reduce the average latency or energy consumed per input.

Table \ref{tab:adaptive_models} reports the details of the models used as a starting point to construct dynamic ensembles, i.e., the rightmost models of the static Pareto curves of Fig.~\ref{fig:results-adaptive-branches}. 
For each model, we report the maximum depth of the trees, the number of estimators, the average number of visited nodes on the test set, the quantization bit-width used for inputs/thresholds ($B_{input}$) and leaves probabilities ($B_{leaves}$), the score (Bal. Accuracy or F1), and the memory occupation.
For Ninapro, we report the average results over the 27 subjects, with the standard deviation in square brackets.
Note that the best score is achieved with different depths, numbers of trees, and quantization precisions for different tasks and ensemble types, demonstrating that all parameters explored during the grid search are critical.

Figure \ref{fig:results-adaptive-branches} compares eight different families of models: on the top row, we compare static GBTs (blue curve) with 6 different adaptive policies, while on the bottom one, we repeat the comparisons for RFs. All the adaptive Pareto curves are obtained by applying an early-stopping policy on top of the ``seed'' models from Table~\ref{tab:adaptive_models}. All points come from the same seed, simply changing the early stopping threshold $t_h$ (whereas, for static models, each point is an entirely different RF/GBT model).
We report the results of five different dynamic inference policies. Namely, we consider the state-of-the-art Max and Score Margin scores from Section~\ref{subsec:methodology_adaptive} in their native form, which uses only the probabilities of the latest executed classifier ($s^t$ and $sm^t$, labelled ``Max'' and ``Score-Margin'' respectively) and in our proposed aggregated variants ($S^t$ and $SM^t$, labelled ``Agg. Max'' and ``Agg. Score-Margin''). Further, we also consider the state-of-the-art QWYC adaptive policy~\cite{qwyc}, which, however, only applies to the binary hard-drive failure detection task. In these experiments, we do not consider batching yet.

On the Ninapro dataset, with dynamic GBTs using our proposed $SM^t$ policy, we are able to consistently reduce the number of visited nodes with respect to static models achieving the same score. The maximum reduction occurs at 71\% (65\%) balanced accuracy for S1 (S2), respectively, where we reduce the number of visited nodes by 54\% (51\%). Conversely, the state-of-the-art adaptive policies fail to achieve the same score, leading to a reduction in accuracy of 9\% (14\%).
Dynamic RFs with $SM^t$, instead, obtain their maximum reduction at 73\% (74\%) balanced accuracy, cutting the number of visited nodes of 83\% (45\%) on the two displayed subjects. Also in this case, the best pre-existing policy, the Score-Margin, obtains a very low accuracy of 59\% (56\%).
Over all subjects in the dataset, we achieve an average maximum reduction of 58.5 $[\pm9]$\% with GBTs and 58.8 [$\pm1.2$]\% with RFs with respect to static models at iso-score.

On the Backblaze dataset, the maximum gain is $88\%$ for GBTs and $69\%$ for RFs, obtained at 66\% and 73\% F1 score. In this case, the QWYC approach is the best one, given its double threshold mechanism, which increases its accuracy when a low number of DTs is employed.
On the other hand, the other pre-existing policy (the Max) leads to significant score drops, respectively of 6\% and 22\% with respect to the seed ensemble.
Lastly, for UniMiB-SHAR, we reduce the number of visited nodes compared to an equally accurate static model by up to $58\%$ and $41\%$ for GBTs and RFs, respectively, at 63\% and 66\% balanced accuracy scores, outperforming the best existing adaptive policy (the Score-Margin), which achieves a maximum accuracy of 55\% and 52\%.

Besides the aforementioned savings, an additional key benefit of dynamic solutions, compared to static models, is their flexibility. In fact, the entire Pareto frontiers of Figure~\ref{fig:results-adaptive-branches} can be obtained by deploying only the seed model and then changing the value of $t_h$ (e.g., depending on battery state or another external trigger). 
Conversely, the static curve is composed of tens of different models, each with different hyperparameters, which can not be simultaneously deployed on the target device due to memory constraints, thus limiting the choices available at runtime.

In Table \ref{tables:hw_unaware}, we compare the static baseline models (``S'' column) and the best dynamic configurations built on top of them which are able to maintain the same score metric (``A-Iso''), or achieve a $< 1\%$ score drop (``A-1\%''). Note that these models are using, for each input, a subset of the DTs included in ``S''.
At iso-score, for the two ensemble types, we achieve a reduction in terms of visted nodes of up to $49\%$ for Ninapro, $88\%$ for Backblaze and $41\%$ for UniMiB-SHAR. If we allow a $1\%$ score drop, the savings increase to up to $70\%$ for Ninapro, $89\%$ for Backblaze and $57\%$ for UniMiB-SHAR.

\begin{table}[ht]
\centering
\begin{tabular}{l|l|lll}
\textbf{Dataset} & \textbf{Model} & \textbf{\#VisitedNodes} & \multicolumn{1}{l}{\textbf{\#Estimators}} & \multicolumn{1}{l}{\textbf{Policy}} \\ \hline \hline
\multicolumn{5}{c}{\textbf{GBTs}}                                                     \\\hline
\multirow{3}{*}{Ninapro}     & S.      & 3060[387] & 37[3.5]   &          \\
                             & A.-Iso & 1805[315] & 22 [3.73] & Agg.SM   \\
                             & A.-1\%       & 1096[191] & 13.4[2.6] & Agg.SM   \\\hline
\multirow{3}{*}{Backblaze}   & S.      & 128       & 9         &          \\
                             & A.-Iso & 14.89     & 1.01      & QWYC o. \\
                             & A.-1\%       & 14.75     & 1.003     & QWYC o. \\\hline
\multirow{3}{*}{UniMiB} & S.      & 2987      & 22        &          \\
                             & A.-Iso & 1766      & 13.02     & Agg.SM   \\
                             & A.-1\%       & 1286      & 9.48      & Agg.SM   \\\hline
\multicolumn{5}{c}{\textbf{RFs}}                                                      \\\hline
\multirow{3}{*}{Ninapro}     & S.      & 348[81]   & 31.8[7]   &          \\
                             & A.-Iso & 175[49]   & 15[3.9]   & Agg.SM   \\
                             & A.-1\%       & 104[30]   & 9[2]      & Agg.SM   \\\hline
\multirow{3}{*}{Backblaze}   & S.      & 156       & 9         &          \\
                             & A.-Iso & 55        & 3.03      & Agg.Max  \\
                             & A.-1\%       & 17        & 1.0005    & QWYC o. \\\hline
\multirow{3}{*}{UniMiB} & S.      & 136       & 10        &          \\
                             & A.-Iso & 116       & 8.46      & Agg.SM   \\
                             & A.-1\%       & 73        & 5.37      & Agg.SM  \\\hline
\end{tabular}
\caption{Statistics of dynamic models compared to their seeds at iso-score (A.-Iso) and with a loss of 1\% accuracy (A.-1\%). Abbreviations: o.: ordered.}\label{tables:hw_unaware}
\end{table}

We notice that in all multi-class classification tasks, the best-performing policy is the proposed aggregated score margin ($SM^t$). On the other hand, on the binary hard-drive failure prediction task, where the $SM^t$ degenerates in the Agg. Max ($S^t$), the QWYC\cite{qwyc} algorithm with ordering works best for 3 out of 4 cases, except for the iso-score RF, which uses $S^t$.
The reason for this is two-fold: first, QWYC uses two separate confidence thresholds for the positive and negative classes, which allows it to execute less DTs on average when predicting that a sample belongs to the ``easiest'' class, i.e., no-failure in this case.
Second, for a binary problem, the Agg. Max and Agg. SM policies become equivalent, as detailed in Section~\ref{subsec:methodology_adaptive}, but the former requires fewer operations, thus obtaining superior trade-offs.
Also, notice that the Max and Score Margin are not present in this table, given that they always fail to reach the same level of accuracy of static models and are outperformed by more than 10\% by our dynamic policies.
In fact, both approaches are tailored for a cascade of increasingly accurate classifiers, which is not the case for tree ensembles of randomly generated week classifiers. 
Therefore, being always sub-optimal compared to our new proposed adaptive policies or to the QWYC algorithm, in the rest of the work, we removed them from the discussion, and we do not consider them for deployment.

\subsection{Dynamic Inference: Tree ordering}
\label{sec:ordering}
In this section, we investigate the impact of the execution order of estimators in dynamic ensembles. The intuitive assumption is that executing the decision trees (DTs) with the highest accuracy first would lead to quicker activation of early stopping policies without affecting accuracy. However, determining the order of trees based on accuracy is not straightforward. %
For example, Figure~\ref{fig:results-adaptive-branches} demonstrates that the performance of the QWYC-ordered ensemble is inferior to that of the QWYC-unordered ensemble. This indicates that the trees achieving the best validation accuracy differ from those maximizing accuracy on the test data.

\begin{figure}[ht]
\centering
\includegraphics[width=0.9\linewidth]{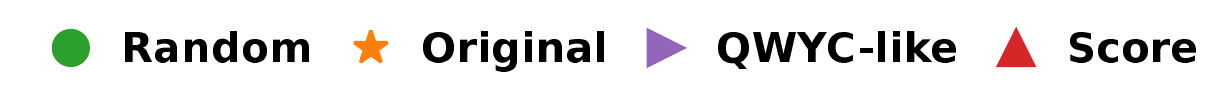}
\includegraphics[width=0.9\linewidth]{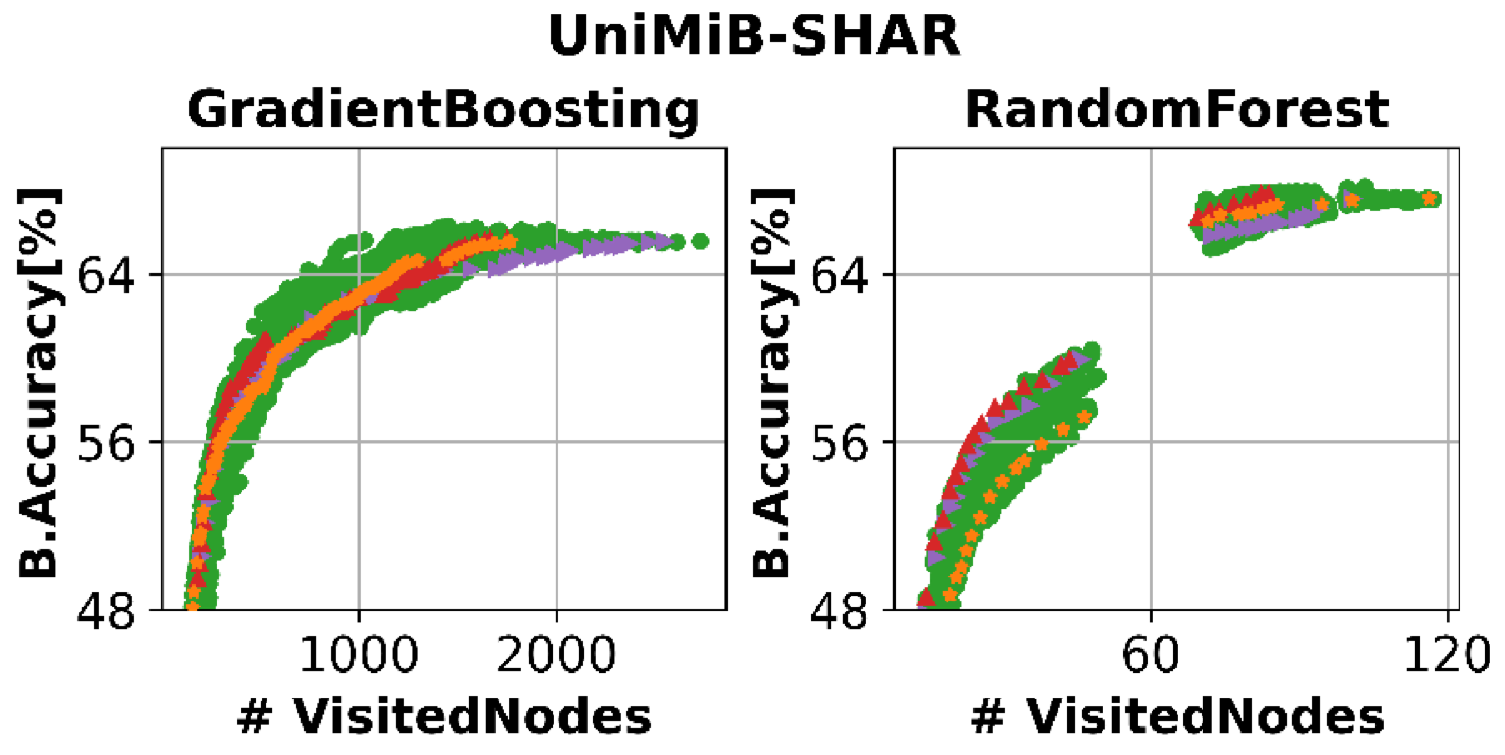}
\caption{Example of dynamic ensembles with different execution orders of the estimators.}
\label{fig:results-ordering}
\end{figure}

Nonetheless, we tested if ordering could improve performance for our new policies. Figure~\ref{fig:results-ordering} shows an example of the results with the Agg. Score Margin policy, on the UniMiB-SHAR dataset (corresponding to the purple markers in the rightmost panels of Fig. \ref{fig:results-adaptive-branches}). We consider 53 different orderings, including: i) 50 randomly generated permutations, ii) two greedy ordering algorithms (QWYC-like and Score), and iii) the original training order.
The QWYC-like order is inspired by~\cite{qwyc}, sorting the estimators in a way that minimizes the number of visited nodes needed to reach iso-accuracy with the static ensemble. The Score order sorts estimators in descending order of accuracy on the validation set.
Each curve corresponds to a different ordering of the \textit{same DTs}, and the different points are generated varying the early-exit threshold. As shown, none of the proposed ``smart'' orders outperform the randomly generated ones, and the original training order falls in the middle of the multiple random curves.

However, selecting the best of the 50 random curves is impossible in practice, because we verified that there is no correlation between the best ordering on the validation set, and the best one on the test set. Similar results are also obtained for other benchmarks and policies, although we omit them for sake of space. Therefore, we conclude that ordering dynamic ensembles based on their performance on the validation set is not a sufficiently robust approach for our benchmarks, and use the natural training order for the rest of our experiments.

\subsection{Dynamic Inference: Deployment Results}
\label{sec:deployment}

\begin{figure*}[ht]
\centering
\includegraphics[width=0.9\linewidth]{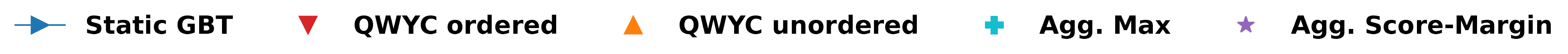}
\includegraphics[width=0.9\linewidth]{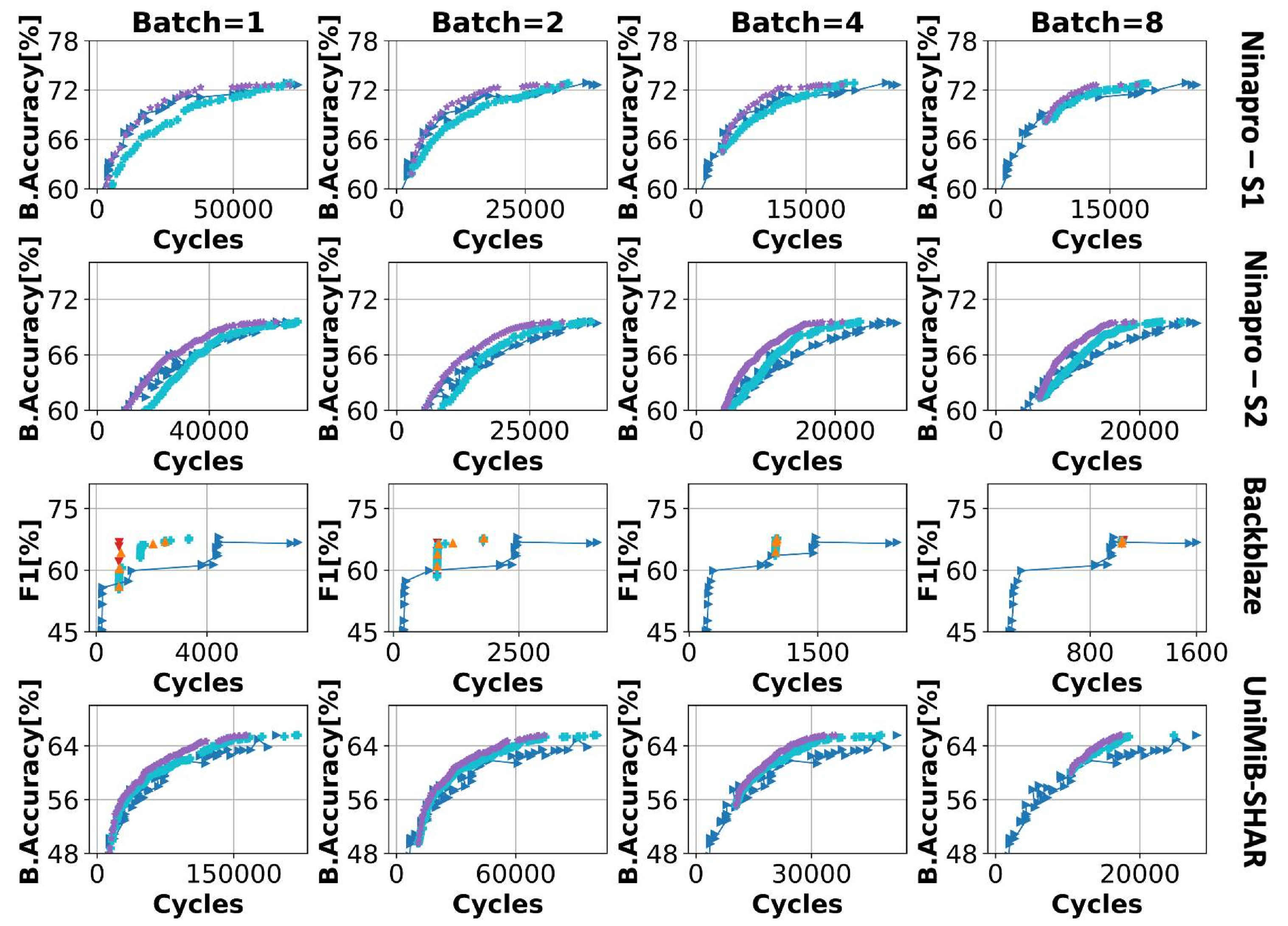}%
\caption{Static and dynamic GBTs Pareto fronts obtained from the validation set and scored on the test set on GAP8. Each column shows a different batch size.}
\label{fig:results-adaptive-cycles-gbt}
\end{figure*}

\begin{figure*}[ht]
\centering
\includegraphics[width=0.9\linewidth]{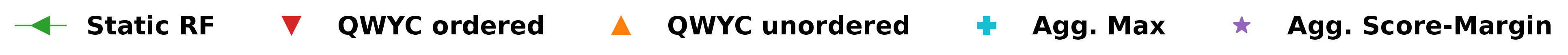}
\includegraphics[width=0.9\linewidth]{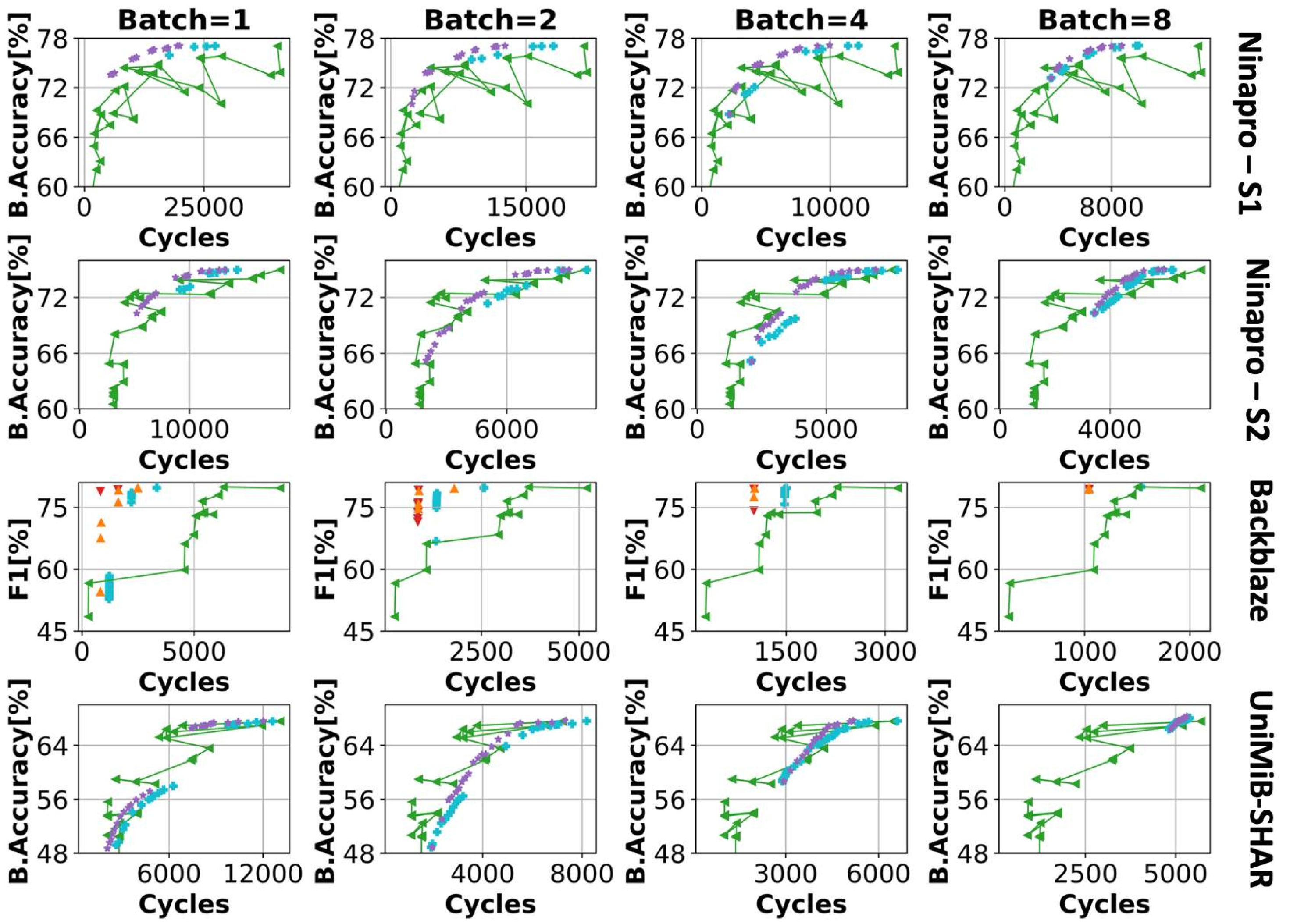}%
\caption{Static and dynamic RFs Pareto fronts obtained from the validation set and scored on the test set on GAP8. Each column shows a different batch size.}
\label{fig:results-adaptive-cycles-rf}
\end{figure*}

Figures \ref{fig:results-adaptive-cycles-gbt} and \ref{fig:results-adaptive-cycles-rf} show static Pareto-optimal ensembles and the dynamic model from Figure~\ref{fig:results-adaptive-branches} when deployed on GAP8. Specifically, we replace the average number of visited nodes with the average number of clock cycles per inference on the target, which correlates with both latency and energy consumption.
In this case, we report results with batch sizes $B$ = 1, 2, 4 and 8. For each value of $B$, we limit the number of cores used to parallelize the execution to $C = B$ for both static and dynamic models, for the reasons explained in Section~\ref{sec:multicore}. The early-exit policy is evaluated after each batch.

Moving from the previous complexity estimate to the actual clock cycles reveals a small advantage of GBTs. For these models, the accumulation of DT's scores on the shared output vector is faster, since each tree only produces a scalar versus a full array of probabilities of $M$ values for RFs (in RFs, a single DT produces $M$ different class probabilities). Given that accumulation happens in a critical section, we find that low-score GBTs outperform low-score RFs on our target, achieving the same score with lower cycles, contrary to the estimate of Figure~\ref{fig:results-adaptive-branches}.
Nonetheless, the general trend is maintained, with RFs rapidly becoming superior as scores increase.

For batch sizes up to $B=4$, dynamic models consistently outperform static solutions for a big portion of the Pareto curve. 
In fact, with less parallelization, the overhead of the early-stopping mechanism is low w.r.t the execution of the static model, leading to large savings.
At $B = 4$, on the Ninapro dataset, we obtain the maximum cycles reduction at 73.9\% (74\%) balanced accuracy for S1 (S2), respectively. With a dynamic RF exploiting the aggregated score margin ($SM^t$) policy, we save 71.8\% (27.1\%) of the cycles compared to the static RF at iso-score.
On the Backblaze dataset, the maximum gain is instead obtained with a GBT at $66.8\%$ F1 score, saving 36.6\% of the cycles.
Lastly, on UniMiB-SHAR, an adaptive GBT reaches 63.4\% balanced accuracy, with 47.7\% fewer cycles compared to the static GBT achieving the same score.

On the contrary, with $B=8$, the introduced overhead becomes significant w.r.t the fast and highly-parallel execution of the ensemble.
In this case, only a reduced set of adaptive models are Pareto optimal. 
Thus, a general conclusion is that \textit{the effectiveness of dynamic early-stopping ensembles reduces with the available cores}.
However, compared to the \textit{most accurate} static models, we still obtain large cycle reductions without loss of accuracy even at $B=8$.

\begin{table*}[]
\resizebox{\textwidth}{!}{
\begin{tabular}{lllllllllllll}
                           & \multicolumn{1}{l|}{}       & \multicolumn{4}{c|}{1 CORE}                                    & \multicolumn{6}{c|}{8 CORES}                                                                                       &          \\\hline
Dataset                    & \multicolumn{1}{l|}{Model}  & Trees C. & Acc. C. & Policy C. & \multicolumn{1}{l|}{Total C.} & Trees C.              & Acc. C. & Policy C.           & Total C.              & E.    & \multicolumn{1}{l|}{Lat.}  & Policy   \\\hline
\multicolumn{13}{c}{\textbf{GBTs}}                                                                                                                                                                                                                        \\\hline
\multirow{3}{*}{Ninapro}   & \multicolumn{1}{l|}{S.}     & 169005   & 9324    & n.a.      & \multicolumn{1}{l|}{178329}   & 21330 (7.92$\times$)  & 9324    & n.a.                & 30654 (5.81$\times$) & 15.63 & \multicolumn{1}{l|}{30.65} &          \\
                           & \multicolumn{1}{l|}{A.-Iso} & 86878    & 7700    & 1496      & \multicolumn{1}{l|}{94578}    & 11950 (7.27$\times$) & 7700    & 204 (7.33$\times$) & 19854 (4.76$\times$) & 10.13 & \multicolumn{1}{l|}{19.85} & Agg. SM  \\
                           & \multicolumn{1}{l|}{A.-1\%}  & 52233    & 4690    & 911       & \multicolumn{1}{l|}{56923}    & 8381 (6.23$\times$)  & 4690    & 136 (6.69$\times$) & 13207 (4.31$\times$) & 6.73  & \multicolumn{1}{l|}{13.21} & Agg. SM  \\\hline
\multirow{3}{*}{Backblaze} & \multicolumn{1}{l|}{S.}     & 7105     & 162     & n.a.      & \multicolumn{1}{l|}{7267}     & 1436 (4.95$\times$)  & 162     & n.a.                & 1598  (4.54$\times$)  & 0.81  & \multicolumn{1}{l|}{1.60}  &          \\
                           & \multicolumn{1}{l|}{A.-Iso} & 4624     & 25      & 50        & \multicolumn{1}{l|}{4699}     & 985 (4.69$\times$)   & 25      & 25 (2.0$\times$)   & 1035 (4.54$\times$)  & 0.53  & \multicolumn{1}{l|}{1.04}  & Agg. Max \\
                           & \multicolumn{1}{l|}{A.-1\%}  & 4624     & 25      & 50        & \multicolumn{1}{l|}{4699}     & 985 (4.69$\times$)   & 25      & 25 (2.0$\times$)   & 1035 (4.54$\times$)  & 0.53  & \multicolumn{1}{l|}{1.04}  & Agg. Max \\\hline
\multirow{3}{*}{UniMiB}    & \multicolumn{1}{l|}{S.}     & 193868   & 2992    & n.a.      & \multicolumn{1}{l|}{196860}   & 24844 (7.80$\times$) & 2992    & n.a.                & 27836 (7.07$\times$) & 14.20 & \multicolumn{1}{l|}{27.84} &          \\
                           & \multicolumn{1}{l|}{A.-Iso} & 75324    & 5533    & 1250      & \multicolumn{1}{l|}{82107}    & 11573 (6.51$\times$) & 5533    & 192 (6.51$\times$) & 17298 (4.74$\times$) & 8.82  & \multicolumn{1}{l|}{17.30} & Agg. SM  \\
                           & \multicolumn{1}{l|}{A.-1\%}  & 52106    & 4029    & 910       & \multicolumn{1}{l|}{57045}    & 10685 (4.88$\times$) & 4029    & 192 (4.73$\times$) & 14906 (3.83$\times$) & 7.60  & \multicolumn{1}{l|}{14.91} & Agg. SM  \\\hline
\multicolumn{13}{c}{\textbf{RFs}}                                                                                                                                                                                                                         \\\hline
\multirow{3}{*}{Ninapro}   & \multicolumn{1}{l|}{S.}     & 22055    & 6720    & n.a.      & \multicolumn{1}{l|}{28775}    & 2892 (7.62$\times$)  & 6720    & n.a.                & 9612 (2.99$\times$)  & 4.90  & \multicolumn{1}{l|}{9.61}  &          \\
                           & \multicolumn{1}{l|}{A.-Iso} & 12580    & 3150    & 1020      & \multicolumn{1}{l|}{16750}    & 2316 (5.43$\times$)  & 3150    & 136 (7.5$\times$)  & 5602 (2.99$\times$)  & 2.86  & \multicolumn{1}{l|}{5.60}  & Agg. SM  \\
                           & \multicolumn{1}{l|}{A.-1\%}  & 9007     & 1890    & 612       & \multicolumn{1}{l|}{11509}    & 1823 (4.94$\times$)  & 1890    & 136 (4.5$\times$)  & 3849 (2.99$\times$)  & 1.96  & \multicolumn{1}{l|}{3.85}  & Agg. SM  \\\hline
\multirow{3}{*}{Backblaze} & \multicolumn{1}{l|}{S.}     & 8676     & 162     & n.a.      & \multicolumn{1}{l|}{8838}     & 1941 (4.47$\times$)  & 162     & n.a.                & 2103 (4.20$\times$)  & 1.07  & \multicolumn{1}{l|}{2.10}  &          \\
                           & \multicolumn{1}{l|}{A.-Iso} & 6289     & 75      & 75        & \multicolumn{1}{l|}{6439}     & 1433 (4.39$\times$)  & 75      & 25 (3.0$\times$)   & 1533 (4.20$\times$)  & 0.78  & \multicolumn{1}{l|}{1.53}  & Agg. Max \\
                           & \multicolumn{1}{l|}{A.-1\%}  & 4300     & 25      & 35        & \multicolumn{1}{l|}{4360}     & 978 (4.40$\times$)   & 25      & 35 (1.0$\times$)   & 1038 (4.20$\times$)  & 0.53  & \multicolumn{1}{l|}{1.04}  & QWYC u.  \\\hline
\multirow{3}{*}{UniMiB}    & \multicolumn{1}{l|}{S.}     & 9400     & 3700    & n.a.      & \multicolumn{1}{l|}{13100}    & 2046 (4.59$\times$)  & 3700    & n.a.                & 5746 (2.28$\times$)  & 2.93  & \multicolumn{1}{l|}{5.75}  &          \\
                           & \multicolumn{1}{l|}{A.-Iso} & 7851     & 3130    & 643       & \multicolumn{1}{l|}{11624}    & 1794 (4.38$\times$)  & 3130    & 152 (4.23$\times$) & 5076 (2.29$\times$)  & 2.59  & \multicolumn{1}{l|}{5.08}  & Agg. SM  \\
                           & \multicolumn{1}{l|}{A.-1\%}  & 8960     & 1986    & 188       & \multicolumn{1}{l|}{11134}    & 2841 (3.15$\times$)  & 1986    & 35 (5.37$\times$)  & 4862 (2.29$\times$)  & 2.48  & \multicolumn{1}{l|}{4.86}  & Agg. Max\\\hline
\end{tabular}
}\caption{Models with $B=1$ and with maximum parallelization ($B=8$) deployed on GAP8. Abbreviations, C.: cycles, Acc.: accumulation, E.: energy, Lat.: latency, u.: unordered.}\label{tab:hw_table}
\end{table*}

Table~\ref{tab:hw_table} reports the cycles, energy, and latency results achieved by the “seed” static models and by two dynamic models, namely the fastest/most efficient ones that achieve the same score, or a score drop of less than 1\%. All models reported refer to the curves with $B = C = 8$.
The table also analyzes in detail the effects of parallelization, providing a breakdown of the cycle counts for the static ``seed'' models and for the various dynamic models, both when
running on 8 cores, and when the same models are executed with $B=C=1$. For each of the 18 ensembles, we report the average cycles for tree execution (Trees C.), probability accumulation (Acc. C.), and policy computation (Policy C.), as well as the total cycles (Total C.). Additionally, for the 8-core case, we also include energy and latency results.
Comparing the $C = 1$ and $C = 8$ configurations, we observe speed-ups ranging from 3.15$\times$ to 7.92$\times$ for tree execution. The suboptimal speed-up is influenced by two factors: the imbalance between trees and the leftover trees executed in the last batch. For example, when executing 9 trees, the first 8 trees are parallelized, while the last one is executed individually, resulting in a maximum speed-up of $\frac{9}{2} = 4.5\times$. It is important to note that only the tree inference section of the ensemble execution is parallelized, as described in Algorithm \ref{alg:dt_inference_deployed} and \ref{alg:adaptive_inference_deployed}.
However, the table also shows a speed-up in the computation of the policy cycles. This is due to the batch size being equal to the number of cores ($B=C$), resulting in the policy being executed $C\times$ fewer times. Also in this case, the speed up is affected by the leftover trees.
The total speed-up on 8 cores ranges from 2.29$\times$ to 7.07$\times$, since it depends both on the parallel tree inference section and on the impact of the sequentially executed accumulation phase.

Overall, when considering 8-core execution we achieve iso-score reductions in terms of latency and energy of up to 41.7\% for Ninapro DB1, 35.2\% for Backblaze, and 37.9\% for UniMiB-SHAR. The maximum gains are obtained by a RF with the $SM^t$ policy for Ninapro DB1, and GBTs with $S^t$ and $SM^t$ policies for Backblaze and UniMiB-SHAR, respectively.
If we allow a score loss of 1\% compared to the seed model, the gains for the three datasets improve to 60\%, 50.6\%, and 46.5\%, respectively.

\section{Conclusions}\label{sec:conclusions}
In this work, we have studied the effectiveness of early-stopping dynamic inference for RFs/GBTs in real-world IoT systems. Namely, thanks to a tool that generates efficient inference code automatically, 
we have deployed optimized static and dynamic tree ensembles, that support parallelization and data quantization, on a multi-core SoC with a complex memory hierarchy.
We benchmarked several adaptive policies, finding that the proposed Aggregated Score Margin obtains the best results for multi-class classification problems, although the improvement with respect to the other proposed approach (Aggregated Max) is often small.
Thanks to the proposed low-cost early stopping policies and batching mechanism, we have shown that we can mitigate the overheads of dynamic inference, which otherwise tend to increase with parallelism. On three IoT-relevant benchmarks, and using all 8 cores available, we have shown that the average energy consumption per inference can be reduced by up to 35.2-41.7\% with respect to a static ensemble, while preserving the same accuracy. Additionally, the obtained dynamic system is extremely flexible, and permits to easily change its working point (in terms of accuracy and energy) by acting on a single tuning parameter.
In our future work, we plan to explore additional lightweight early stopping policies for edge devices, e.g., considering a running mean of scores rather than a simple aggregation, and focus on optimizing the execution of small adaptive tree-based models for even smaller platforms, with tighter memory constraints.

\bibliographystyle{IEEEtran}
\bibliography{IEEEabrv,bibliography/lib}

\begin{IEEEbiography}[{\includegraphics[width=1in,height=1.25in,clip,keepaspectratio]{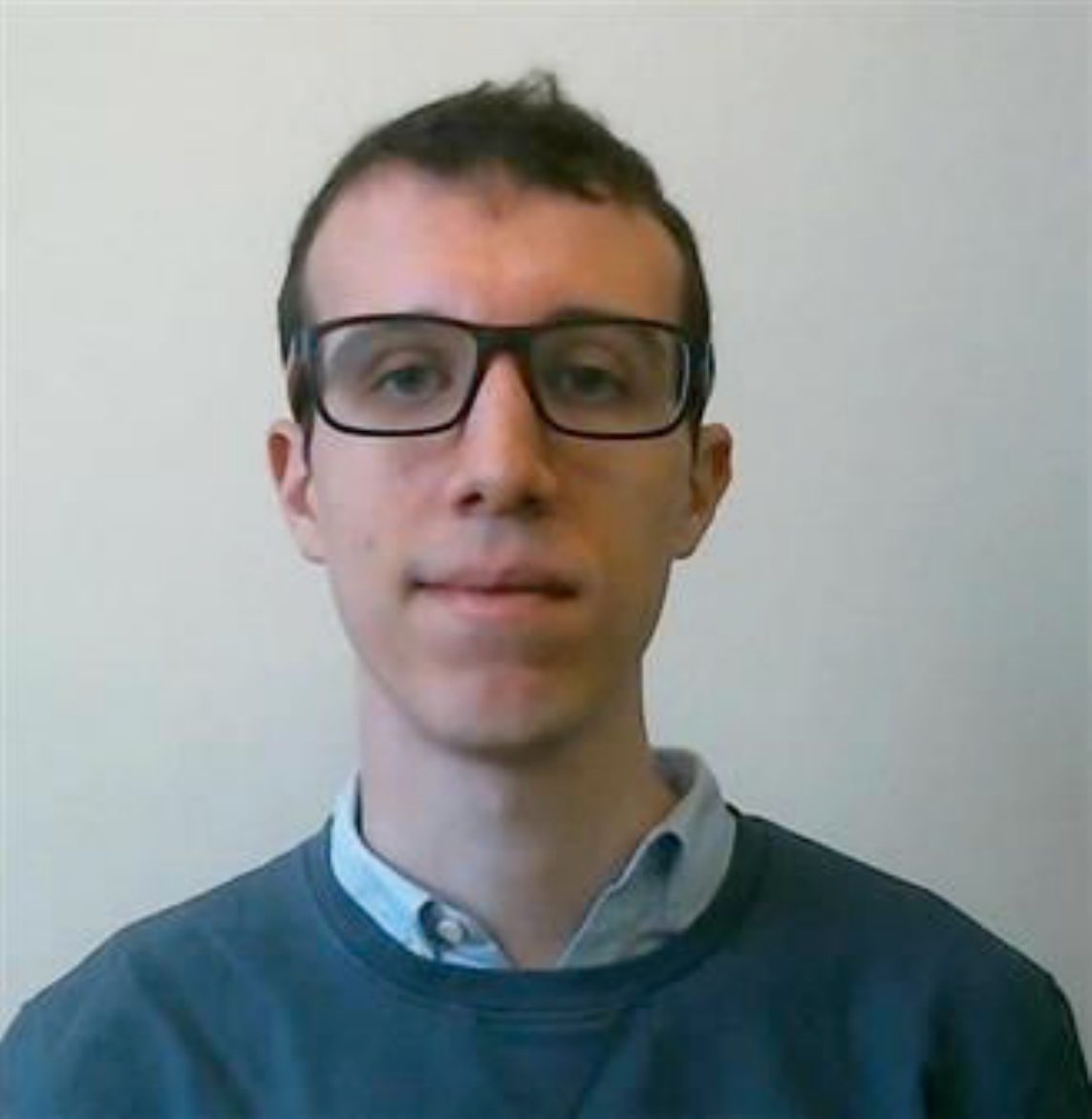}}]{Francesco Daghero} is a PhD student at Politecnico di Torino. He received a M.Sc. degree in computer engineering from Politecnico di Torino, Italy, in 2019. His research interests concern embedded machine learning and Industry 4.0.
\end{IEEEbiography}

\begin{IEEEbiography}[{\includegraphics[width=1in,height=1.25in,clip,keepaspectratio]{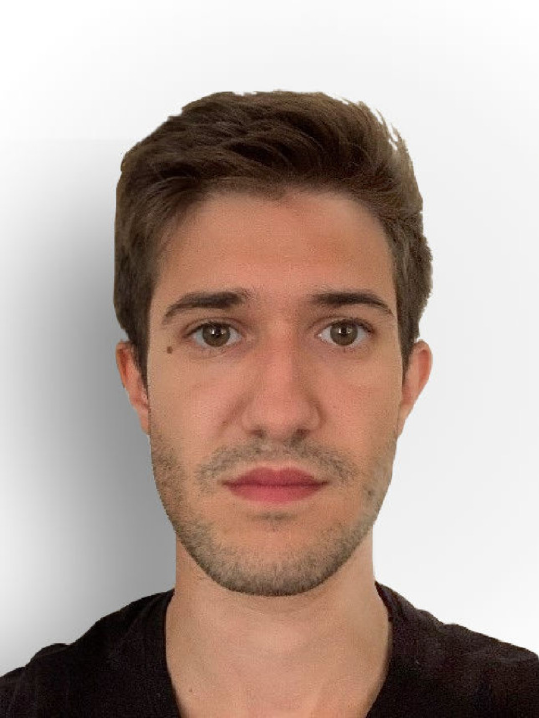}}]{Alessio Burrello}
received his M.Sc. and Ph.D degrees in Electronic Engineering at the Politecnico of Turin, Italy, and University of Bologna, respectively, in 2018 and 2023.
He is currently a research assistant at Politecnico di Torino.
His research interests include parallel programming models for embedded systems, machine and deep learning, hardware-oriented deep learning, and code optimization for multi-core systems.
\end{IEEEbiography}

\begin{IEEEbiography}[{\includegraphics[width=1in,height=1.25in,clip,keepaspectratio]{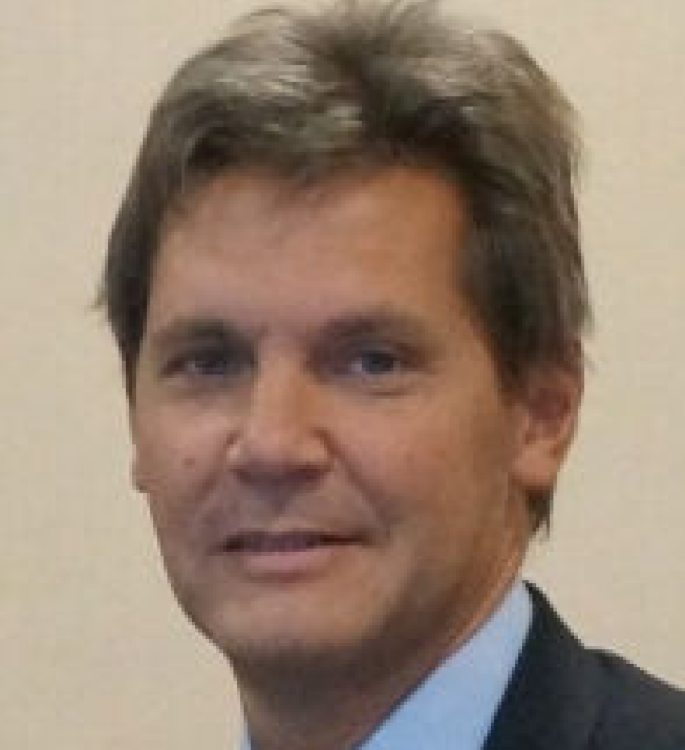}}]{Enrico Macii} is a Full Professor of Computer Engineering with the Politecnico di Torino, Torino, Italy. He holds a Laurea degree in electrical engineering from the Politecnico di Torino, a Laurea degree in computer science from the Universita' di Torino, Turin, and a PhD degree in computer engineering from the Politecnico di Torino. His research interests are in the design of digital electronic circuits and systems, with a particular emphasis on low-power consumption aspects energy efficiency, sustainable urban mobility, clean and intelligent manufacturing. He is a Fellow of the IEEE.
\end{IEEEbiography}

\begin{IEEEbiography}[{\includegraphics[width=1in,height=1.25in,clip,keepaspectratio]{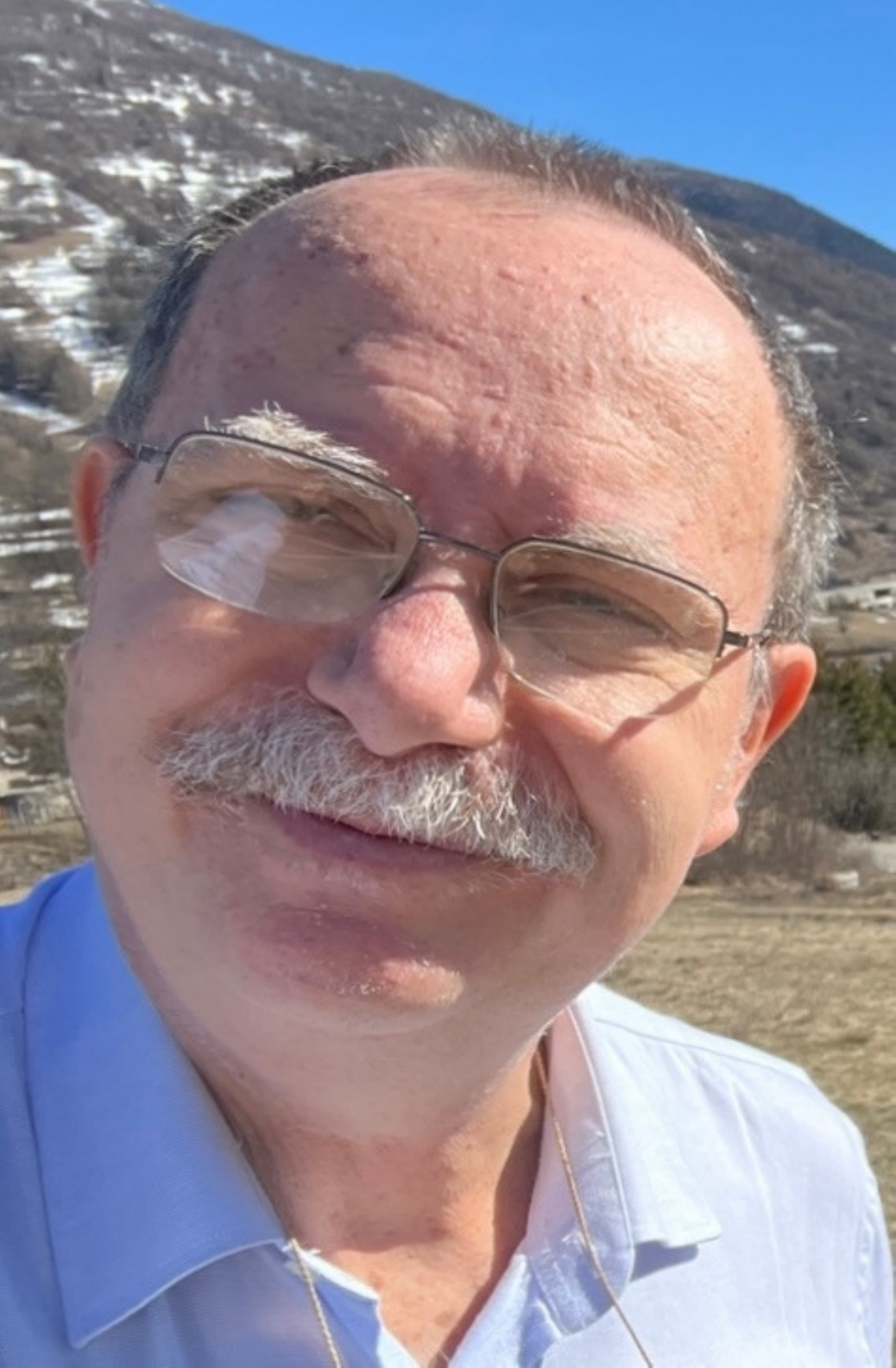}}]{Paolo Montuschi} (M’90-SM’07-F’14) (paolo.montuschi@polito.it) is a full professor with the Department of Control and Computer Engineering, Rector's Delegate for Information Systems, and a past member of the Board of Governors at Politecnico di Torino, Italy. His research interests include computer arithmetic, computer architectures, and intelligent systems. He is an IEEE Fellow, a life member of the International Academy of Sciences in Turin, and of HKN, the Honor Society of IEEE. He serves as the Editor-in-Chief of the IEEE Transactions on Emerging Topics in Computing, the 2020-23 Chair of the IEEE TAB/ARC, and a member of the IEEE Awards Board. Previously, he served in a number of positions, including the Editor-in-Chief of the IEEE Transactions on Computers (2015-18), the 2017-20 IEEE Computer Society Awards Committee Chair, a Member-at-Large of IEEE PSPB (2015-20), and as the Chair 
of its Strategic Planning Committee (2019-20).
More information at \url{ http://staff.polito.it/paolo.montuschi} \end{IEEEbiography}

\begin{IEEEbiography}[{\includegraphics[width=1in,height=1.25in,clip,keepaspectratio]{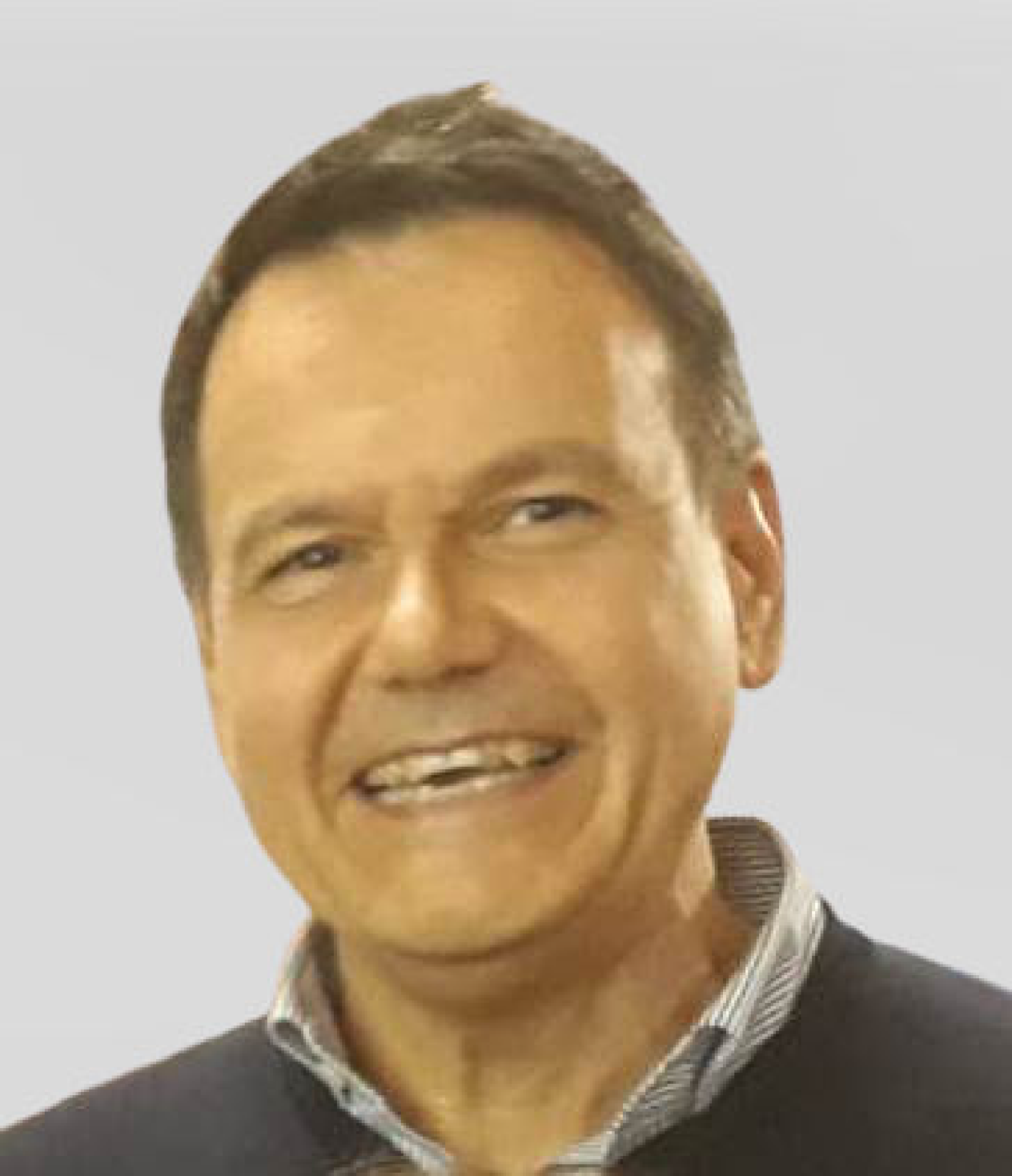}}]{Massimo Poncino} is a Full Professor of Computer Engineering with the Politecnico di Torino, Torino, Italy. His current research interests include various aspects of design automation of digital systems, with emphasis on the modeling and optimization of energy-efficient systems. He received a PhD in computer engineering and a Dr.Eng. in electrical engineering from Politecnico di Torino. He is a Fellow of the IEEE.
\end{IEEEbiography}

\begin{IEEEbiography}[{\includegraphics[width=1in,height=1.25in,clip,keepaspectratio]{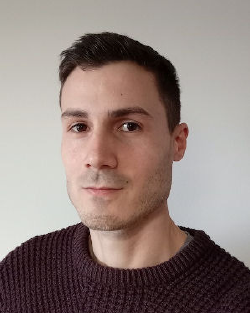}}]{Daniele Jahier Pagliari} received the M.Sc. and Ph.D. degrees in computer engineering from the Politecnico di Torino, Turin, Italy, in 2014 and 2018, respectively. He is currently an Assistant Professor with the Politecnico di Torino. His research interests are in the computer-aided design and optimization of digital circuits and systems, with a particular focus on energy-efficiency aspects and on emerging applications, such as machine learning at the edge.
\end{IEEEbiography}

\end{document}